\documentclass[12pt]{article}

\newcommand{\blind}{0}  

\usepackage{graphics,latexsym,amsthm,epsfig,bm,fullpage,setspace,amsmath,textcomp,url,xcolor}
\usepackage[pagewise,right]{lineno}
\usepackage{natbib}
\usepackage{dsfont}
\usepackage{amssymb}

\def\argmin{\mathop{\rm argmin}}

\newcommand{\ind}[1]{\mathds{1}{\{ {#1} \} }}
\newcommand{\bw}{\boldsymbol{w}}

\newtheorem{theorem}{Theorem}
\newtheorem{lemma}{Lemma}
\newtheorem{corollary}{Corollary}
\newtheorem{pro}{Proposition}
\newtheorem{defi}{Definition}
\newtheorem{remark}{Remark}

\def\mydef{\,{\buildrel \Delta \over =}\,}

\usepackage[top=1in, bottom=1in, left=1in, right=1in]{geometry}
\begin{document}
\pagenumbering{roman}
\def\spacingset#1{\renewcommand{\baselinestretch}%
{#1}\small\normalsize} \spacingset{1}

\if0\blind
{
  \title{\bf Stabilized Nearest Neighbor Classifier and Its Statistical Properties}
  \author{
    Wei Sun\\Yahoo Labs\\ \\and \\
    Xingye Qiao\\Department of Mathematical Sciences\\Binghamton University, State University of New York\\ \\and\\
    Guang Cheng\thanks{Correspondence to Guang Cheng (e-mail: chengg@purdue.edu). Wei Sun (e-mail: sunweisurrey@yahoo-inc.com) is a Research Scientist at Yahoo! Labs, 701 First Avenue, Sunnyvale, CA 94089; Xingye Qiao (e-mail: qiao@math.binghamton.edu) is an Assistant Professor at Department of Mathematical Sciences at Binghamton University, State University of New York, Binghamton, NY 13902-6000; and Guang Cheng (e-mail: chengg@purdue.edu) is an Associate Professor at Department of Statistics at Purdue University, West Lafayette, IN 47906. Qiao's research is partially supported by Simons Foundation \#246649. Cheng's research is partially supported by NSF CAREER Award DMS-1151692, DMS-1418042, Simons Fellowship in Mathematics, Office of Naval Research (ONR N00014-15-1-2331) and a grant from Indiana Clinical and Translational Sciences Institute. All authors gratefully acknowledges Statistical and Applied Mathematical Sciences Institute (SAMSI) for the hospitality and support during their visits when this work was initiated. Guang Cheng acknowledges Department of Operations Research and Financial Engineering at Princeton University for the hospitality and support during his sabbatical when the revision of this work was conducted.  The authors thank Professor Richard Samworth for personal communications and thank the editors, the associate editor, and three anonymous reviewers for their helpful comments and suggestions which led to a much improved presentation. This research work was conducted when Wei Sun was a Ph.D. candidate at Department of Statistics at Purdue University, West Lafayette, IN 47906.}\\Department of Statistics\\
Purdue University}
  \date{}
  \maketitle
} \fi

\if1\blind
{
  \bigskip
  \bigskip
  \bigskip
  \begin{center}
    {\LARGE\bf Stabilized Nearest Neighbor Classifier and Its Statistical Properties}
  \end{center}
  \medskip
} \fi

\begin{abstract}
The stability of statistical analysis is an important indicator for reproducibility, which is one main principle of scientific method. It entails that similar statistical conclusions can be reached based on independent samples from the same underlying population. In this paper, we introduce a general measure of classification instability (CIS) to quantify the sampling variability of the prediction made by a classification method. Interestingly, the asymptotic CIS of any weighted nearest neighbor classifier turns out to be proportional to the Euclidean norm of its weight vector. Based on this concise form, we propose a stabilized nearest neighbor (SNN) classifier, which distinguishes itself from other nearest neighbor classifiers, by taking the stability into consideration. In theory, we prove that SNN attains the minimax optimal convergence rate in risk, and a sharp convergence rate in CIS. The latter rate result is established for general plug-in classifiers under a low-noise condition. Extensive simulated and real examples demonstrate that SNN achieves a considerable improvement in CIS over existing nearest neighbor classifiers, with comparable classification accuracy. We implement the algorithm in a publicly available R package \texttt{snn}.
\end{abstract}

\vfill
\newpage

\spacingset{1.45} 

\pagenumbering{arabic}
\setcounter{page}{1}

\section{Introduction}
Data science has become a driving force for many scientific studies. As datasets get bigger and the methods of analysis become more complex, the need for reproducibility has increased significantly \citep{SLP14}. A minimal requirement of reproducibility is that one can reach similar results based on independently generated datasets. The issue of reproducibility has drawn much attention in the scientific community (see a special issue of \textit{Nature}\footnote{at http://www.nature.com/nature/focus/reproducibility/}); Marcia McNutt, the Editor-in-Chief of \textit{Science}, pointed out that ``reproducing an experiment is one important approach that scientists use to gain confidence in their conclusions.'' In other words, if conclusions cannot be reproduced, the credit of the researchers, along with the scientific conclusions themselves, will be in jeopardy.

\subsection{Stability}
Statistics as a subject can help improve reproducibility in many ways. One particular aspect we stress in this article is the stability of a statistical procedure used in the analysis. According to \cite{Y13}, ``reproducibility manifests itself in stability of statistical results relative to `reasonable' perturbations to data and to the model used.'' An instable statistical method leads to the possibility that a correct scientific conclusion is not reproducible, and hence is not recognized, or even falsely discredited. 

Stability has indeed received much attention in statistics. However, few work has focused on stability itself. Many works instead view stability as a tool for other purposes. For example, in clustering problems, \citet{BEG02} introduced the clustering instability to assess the quality of a clustering algorithm; \citet{W10} used the clustering instability as a criterion to select the number of clusters. In high-dimensional regression, \citet{MB10} proposed stability selection procedures for variable selection; \citet{LRW10} and \citet{SWF13} applied stability for tuning parameter selection. For more applications, see the use of stability in model selection \citep{B96}, analyzing the effect of bagging \citep{BY02}, and deriving the generalization error bound \citep{BE02,EEP05}. While successes of stability have been reported in the aforementioned works, to the best of our knowledge, there has been little systematic methodological and theoretical study of stability itself in the classification context.

On the other hand, we are aware that ``a study can be reproducible but still be wrong''\footnote{http://simplystatistics.org/2014/06/06/the-real-reason-reproducible-research-is-important/}. So can a classification method be stable but inaccurate. Thus, in this article, stability is not meant to replace classification accuracy, which is the primary goal for much of the research work on classification. However, an irreproducible or instable study will definitely reduce its chance of being accepted by the scientific community, no matter how accurate it is. Hence, it is ideal for a method to be both accurate and stable, a goal of the current article.

Moreover, in certain practical domains of classification, stability can be as important as accuracy. This is because providing a stable prediction plays a crucial role on users' trust on a system. For example, Internet streaming service provider Netflix has a movie recommendation system based on complex supervised learning algorithms. In this application, if two consecutively recommended movies are from two totally different genres, the viewers can immediately perceive such instability, and have a bad user experience with the service \citep{AZ10}.

\subsection{Overview}
The $k$-nearest neighbor ($k$NN) classifier \citep{FH51,CH67} is one of the most popular nonparametric classification methods, due to its conceptual simplicity and powerful prediction capability. In the literature, extensive research have been done to justify various nearest neighbor classifiers based on the risk, which measures the inaccuracy of a classifier \citep{DW77, S77, G81, DGKL94, SV98, BCG10}. We refer the readers to \citet{DGL96} for a comprehensive study. Recently, \citet{S12} has proposed an optimal weighted nearest neighbor (OWNN) classifier. Like most other existing nearest neighbor classifiers, OWNN focuses on the risk without paying attention to the classification stability.

\begin{figure}[t!]
\begin{center}
\includegraphics[scale=0.6]{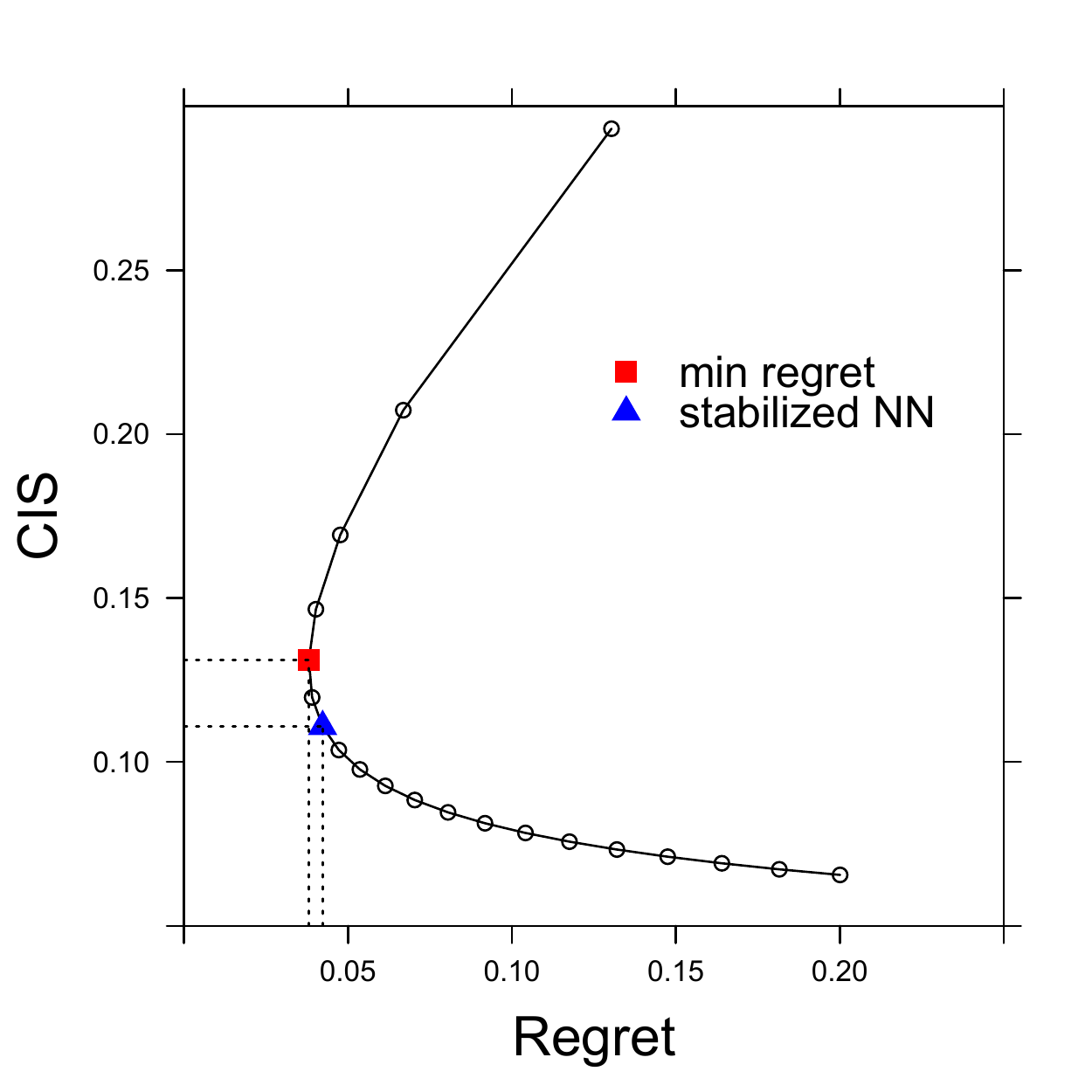}\vspace{-1em}
\caption{\label{stabknn_regret_cis} \footnotesize Regret and CIS of the $k$NN classifier. From top to bottom, each circle represents the $k$NN classifier with $k\in \{1,2,\dots,20\}$. The red square corresponds to the classifier with the minimal regret and the classifier depicted by the blue triangle improves it to have a lower CIS.}
\end{center}
\end{figure}

In this article, we define a general measure of stability for a classification method, named as \textit{Classification Instability} (CIS). It characterizes the sampling variability of the prediction. An important result we show is that the asymptotic CIS of any weighted nearest neighbor classifier (a generalization of  $k$NN), denoted as WNN, turns out to be proportional to the Euclidean norm of its weight vector. This rather concise form is crucial in our methodological development and theoretical analysis. To illustrate the relation between risk and CIS, we apply the $k$NN classifier to a toy example (see details in Section \ref{sec:validation}) and plot in Figure \ref{stabknn_regret_cis} the regret (that is, the risk minus a constant known as the Bayes risk) versus CIS, calculated according to Proposition \ref{thm:regret} and Theorem \ref{thm:CIS} in Section~\ref{sec:snn0}, for different $k$. As $k$ increases, the classifier becomes more and more stable, while the regret first decreases and then increases. In view of the $k$NN classifier with the minimal regret, marked as the red square in Figure \ref{stabknn_regret_cis}, one may have the impression that there are other $k$ values with similar regret but much smaller CIS, such as the one marked as the blue triangle shown in the plot.

Inspired by Figure \ref{stabknn_regret_cis}, we propose a novel method called stabilized nearest neighbor (SNN) classifier, which takes the stability into consideration. The SNN procedure is constructed by minimizing the CIS of WNN over an acceptable region where the regret is small, indexed by a tuning parameter. SNN encompasses the OWNN classifier as a special case.

To understand the theoretical property of SNN, we establish a sharp convergence rate of CIS for general plug-in classifiers. This sharp rate is slower than but approaching $n^{-1}$, shown by adapting the framework of \citet{AT07}. Furthermore, the proposed SNN method is shown to achieve both the minimax optimal rate in the regret established in the literature, and the sharp rate in CIS established in this article.

\begin{figure}[t!]
\begin{center}
\vspace{-1em}\includegraphics[scale=0.6]{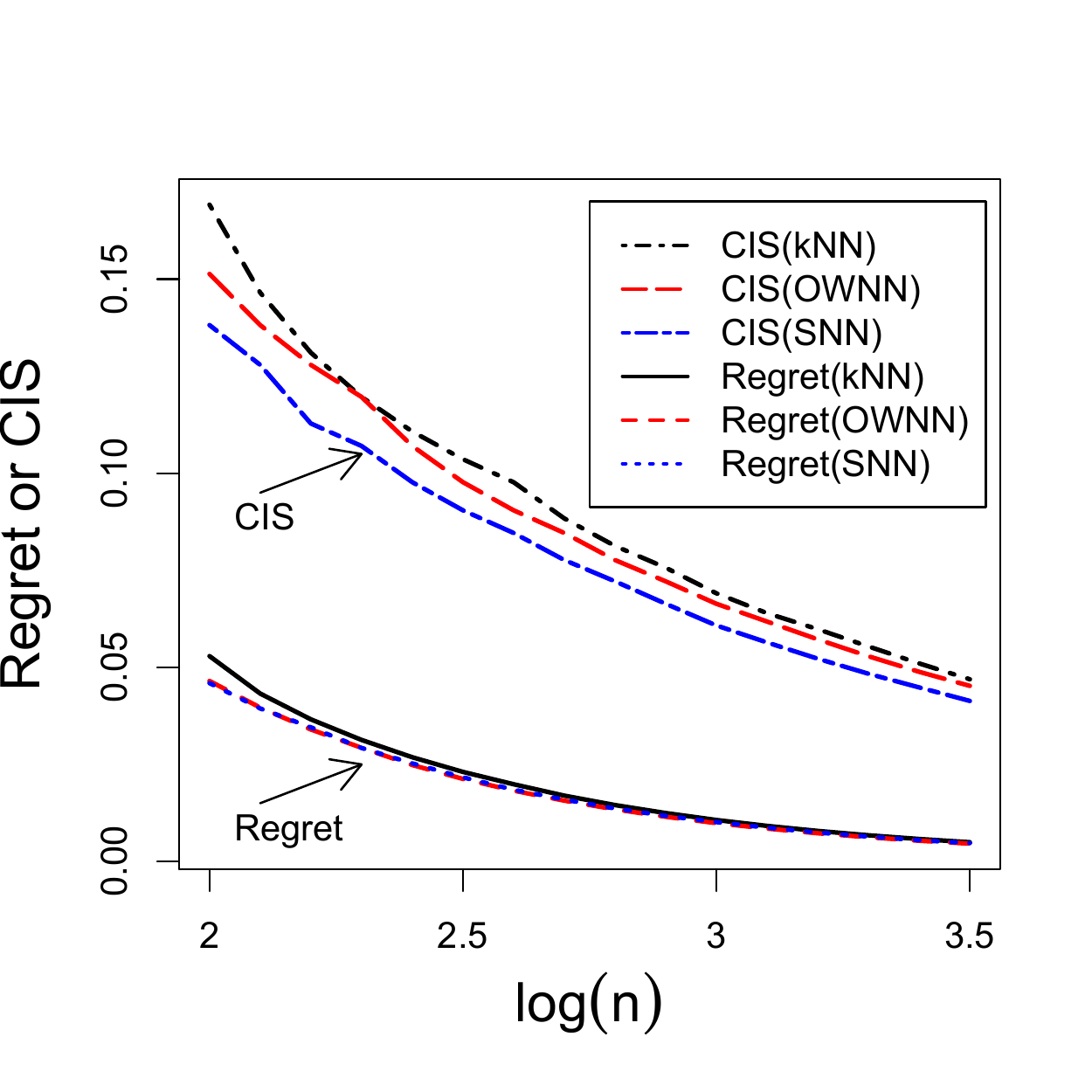}\vspace{-1em}
\caption{\label{regret_cis_comparison} \footnotesize Regret and CIS of $k$NN, OWNN, and SNN procedures for a bivariate normal example. The top three lines represent CIS's of $k$NN, OWNN, and SNN. The bottom three lines represent regrets of $k$NN, SNN, and OWNN. The sample size shown on the x-axis is in the $\log_{10}$ scale.}
\end{center}
\end{figure}

To further illustrate the advantages of the SNN classifier, we offer a comprehensive asymptotic comparison among various classifiers, through which new insights are obtained. It is theoretically verified that the CIS of our SNN procedure is much smaller than those of others. Figure \ref{regret_cis_comparison} shows the regret and CIS of $k$NN, OWNN, and SNN for a bivariate example (see details in Section \ref{sec:validation}). Although OWNN is \textit{theoretically} the best in regret, its regret curve appear to overlap with that of SNN. On the other hand, the SNN procedure has a noticeably smaller CIS than OWNN. A compelling message is that with almost the same accuracy, our SNN could greatly improve stability. In the finite sample case, extensive experiments confirm that SNN has a significant improvement in CIS, and sometimes even improves accuracy slightly. Such appealing results are supported by our theoretical finding (in Corollary \ref{cor:snn_ownn}) that the regret of SNN approaches that of OWNN at a faster rate than the rate at which the CIS of OWNN approaches that of SNN, where both rates are shown to be sharp. As a by-product, we also show that OWNN is more stable than $k$NN and bagged nearest neighbor (BNN) classifiers.

The rest of the article is organized as follows. Section~\ref{sec:cis} defines CIS for a general classification method. In Section~\ref{sec:snn0}, we study the stability of the nearest neighbor classifier, and propose a novel SNN classifier. The SNN classifier is shown to achieve an established sharp rate in CIS and the minimax optimal rate in regret in Section~\ref{sec:thms}. Section~\ref{theoretical_compare} presents a thorough theoretical comparison of regret and CIS between the SNN classifier and other nearest neighbor classifiers. Section \ref{sec:algorithm} discusses the issue of tuning parameter selection, followed by numerical studies in Section~\ref{sec:exp}. We conclude the article in Section~\ref{discussion}. The appendix and supplementary materials are devoted to technical proofs.

\section{Classification Instability}\label{sec:cis}
Let $(X,Y)\in{\mathbb R}^d \times\{1,2\}$ be a random couple with a joint distribution $P$. We regard $X$ as a $d$-dimensional vector of features for an object and $Y$ as a label indicating that the object belongs to one of two classes. Denote the prior class probability  as $\pi_1={\mathbb P}(Y=1)$, where $\mathbb P$ is the probability with respect to $P$, and the distribution of $X$ given $Y=r$ as $P_r$ with $r=1,2$. The marginal distribution of $X$ can be written as $\bar{P}=\pi_1 P_1 + (1-\pi_1)P_2$. For a classifier $\phi: {\mathbb R}^d \mapsto\{1,2\}$, the risk of $\phi$ is defined as $R(\phi) = \mathbb P(\phi(X)\ne Y)$. It is well known that the Bayes rule, denoted as $\phi^{\textrm{Bayes}}$, minimizes the above risk. Specifically, $\phi^{\textrm{Bayes}}(x)=1+\ind{\eta(x)< 1/2}$, where $\eta(x)={\mathbb P}(Y=1|X=x)$ and $\ind{\cdot}$ is the indicator function. In practice, a classification procedure $\Psi$ is applied to a training data set ${\cal D}= \{(X_i,Y_i), i=1,\ldots,n\}$ to produce a classifier $\widehat\phi_{n}=\Psi({\cal D})$. We define the risk of the procedure $\Psi$ as $\mathbb E_{{\cal D}}[R(\widehat\phi_{n})]$, and the regret of $\Psi$ as $\mathbb E_{{\cal D}}[R(\widehat\phi_{n})]-R(\phi^{\textrm{Bayes}})$, where $\mathbb E_{{\cal D}}$ denotes the expectation with respect to the distribution of $\cal D$, and $R(\phi^{\textrm{Bayes}})$ is called Bayes risk. Both the risk and regret describe the inaccuracy of a classification method. In practice, for a classifier $\phi$, the classification error for a test data can be calculated as an empirical version of $R(\phi)$.

For a classification procedure, it is desired that, with high probability, classifiers trained from different samples yield the same prediction for the same object. Our first step in formalizing the classification instability is to define the distance between two generic classifiers $\phi_1$ and $\phi_2$, which measures the level of disagreement between them.
\begin{defi}
(Distance between Classifiers) Define the distance between two classifiers $\phi_1$ and $\phi_2$ as $d(\phi_1,\phi_2) = {\mathbb P}(\phi_1(X) \ne \phi_2(X))$.
\end{defi}
We next define the classification instability (CIS). Throughout the article, we denote ${\cal D}_1$ and ${\cal D}_2$ as two i.i.d. copies of the training sample ${\cal D}$. For ease of notation, we have suppressed the dependence of $\textrm{CIS}(\Psi)$ on the sample size $n$ of $\mathcal D$.
\begin{defi}(Classification Instability)
Define the classification instability of a classification procedure $\Psi$ as
\begin{equation}
\textrm{CIS}(\Psi) = {\mathbb E}_{{\cal D}_1,{\cal D}_2}\Big[ d(\widehat\phi_{n1},\widehat\phi_{n2})\Big]\label{CIS}
\end{equation}
where $\widehat\phi_{n1}=\Psi(\mathcal D_1)$ and $\widehat\phi_{n2}=\Psi(\mathcal D_2)$ are the classifiers obtained by applying the classification procedure $\Psi$ to samples ${\cal D}_1$ and ${\cal D}_2$.
\end{defi}
Intuitively, CIS is an average probability that the same object is classified to two different classes in two separate runs of a learning algorithm. By definition, $0\le \textrm{CIS}(\Psi) \le 1$, and a small $\textrm{CIS}(\Psi)$ represents a stable classification procedure $\Psi$.

\section{Stabilized Nearest Neighbor Classifier}\label{sec:snn0}

\subsection{Review of WNN}\label{sec:WNN}

For any fixed $x$, let $(X_{(1)},Y_{(1)}),\ldots, (X_{(n)},Y_{(n)})$ be a sequence of observations with ascending distance to $x$. For a nonnegative weight vector $\bw_n=(w_{ni})_{i=1}^n$ satisfying $\sum_{i=1}^n w_{ni} = 1$, a WNN classifier $\widehat{\phi}_{n}^{\bw_n}$ predicts the label of $x$ as $\widehat{\phi}_{n}^{\bw_n}(x)=1+\ind{\sum_{i=1}^n w_{ni}\ind{Y_{(i)}=1}< 1/2}$. \citet{S12} revealed a nice asymptotic expansion formula for the regret of WNN.

\begin{pro}
\label{thm:regret}
\citep{S12} Under Assumptions (A1)--(A4) defined in Appendix \ref{sec:assumptions}, for each $\beta\in (0,1/2)$, we have, as $n\rightarrow \infty$,
\begin{equation}
\textrm{Regret}(\textrm{WNN}) = \left\{B_1 \sum_{i=1}^n w_{ni}^2 + B_2 \Big (\sum_{i=1}^n \frac{\alpha_i w_{ni}}{n^{2/d}}\Big)^2 \right\}\{1+o(1)\},\label{regret}
\end{equation}
uniformly for $\bw_n\in W_{n,\beta}$ with $W_{n,\beta}$ defined in Appendix \ref{sec:defwnb}, where $\alpha_i=i^{1+\frac{2}{d}}-(i-1)^{1+\frac{2}{d}}$, and constants $B_1$ and $B_2$ are defined in Appendix \ref{sec:defwnb}.
\end{pro}

\citet{S12} further derived a weight vector that minimizes the asymptotic regret $(\ref{regret})$ which led to the optimal weighted nearest neighbor (OWNN) classifier.

\subsection{Asymptotically Equivalent Formulation of CIS}\label{sec:asy_CIS}

Denote two resulting WNN classifiers trained on ${\cal D}_1$ and ${\cal D}_2$ as $\widehat{\phi}_{n1}^{\bw_n}(x)$ and $\widehat{\phi}_{n2}^{\bw_n}(x)$ respectively. With a slight abuse of notation, we denote the CIS of a WNN classification procedure by $\textrm{CIS}({\makebox{WNN}})$. According to the definition in (\ref{CIS}), classification instability of a WNN procedure is
$\textrm{CIS}(\textrm{WNN})= {\mathbb P}_{{\cal D}_1,{\cal D}_2,X}\Big(\widehat{\phi}_{n1}^{\bw_n}(X) \ne \widehat{\phi}_{n2}^{\bw_n}(X)\Big).$
Theorem \ref{thm:CIS} provides an asymptotic expansion formula for the CIS of WNN in terms of its weight vector $\bw_n$.
\begin{theorem}
\label{thm:CIS}
(Asymptotic CIS) Under Assumptions (A1)--(A4) defined in Appendix \ref{sec:assumptions}, for each $\beta\in (0,1/2)$, we have, as $n\rightarrow \infty$,
\begin{equation}
\textrm{CIS}(\textrm{WNN}) = B_3 \Big(\sum_{i=1}^n w_{ni}^2\Big)^{1/2} \{1+o(1)\}, \label{eq:asy_CIS}
\end{equation}
uniformly for all $\bw_n\in W_{n,\beta}$ with $W_{n,\beta}$ defined in Appendix \ref{sec:defwnb}, where the constant $B_3 = 4B_1/\sqrt{\pi}>0$ with $B_1$ defined in Appendix \ref{sec:defwnb}.
\end{theorem}

Theorem \ref{thm:CIS} demonstrates that the asymptotic CIS of a WNN procedure is proportional to $(\sum_{i=1}^n w_{ni}^2)^{1/2}$. For example, for the $k$NN procedure (that is the WNN procedure with $w_{ni}=k^{-1}\ind{1\le i\le k}$), its CIS is asymptotically $B_3\sqrt{1/k}$. Therefore, a larger value of $k$ leads to a more stable $k$NN procedure, which was seen in Figure \ref{stabknn_regret_cis}. Furthermore, we note that the CIS expansion in (\ref{eq:asy_CIS}) is related to the first term in (\ref{regret}). The expansions in (\ref{regret}) and (\ref{eq:asy_CIS}) allow precise calibration of regret and CIS. This delicate connection is important in the development of our SNN procedure.

\subsection{Stabilized Nearest Neighbor Classifier}
\label{sec:SNN}

To stabilize WNN, we consider a weight vector which minimizes the CIS over an acceptable region where the classification regret is less than some constant $c_1>0$, that is,
\begin{align}
\min_{\bw_n}~~ &\textrm{CIS}(\textrm{WNN}), \label{original}\\
\textrm{subject to}~~ &\textrm{Regret}(\textrm{WNN}) \le c_1,~\sum_{i=1}^n w_{ni} =1,~\bw_{n} \ge 0. \nonumber
\end{align}
By a non-decreasing transformation, we change the objective function in (\ref{original}) to $\textrm{CIS}^2(\textrm{WNN})$. Furthermore, considering the Lagrangian formulation, we can see that (\ref{original}) is equivalent to minimizing $\textrm{Regret}(\textrm{WNN}) + \lambda_0\textrm{CIS}^2(\textrm{WNN})$ subject to the constraints that $\sum_{i=1}^n w_{ni} =1$ and $\bw_{n} \ge 0$, where $\lambda_0>0$. The equivalence is ensured by the expansions (\ref{regret}) and (\ref{eq:asy_CIS}) in Proposition \ref{thm:regret} and Theorem \ref{thm:CIS}, and the fact that both the objective function and the constraints are convex in the variable vector $\bw_n$. The resulting optimization is
\begin{align}
\min_{\bw_n}~~ &\left(\sum_{i=1}^n \frac{\alpha_i w_{ni}}{n^{2/d}}\right)^2 + \lambda \sum_{i=1}^n w_{ni}^2,\label{formal}\\
\textrm{subject to}~~ &\sum_{i=1}^n w_{ni} =1,~\bw_{n} \ge 0, \nonumber
\end{align}
where $\lambda = (B_1+\lambda_0B_3^2)/B_2$ depends on constants $B_1$ and $B_2$ and $\lambda_0$. When $\lambda\rightarrow \infty$, (\ref{formal}) leads to the most stable but trivial $k$NN classifier with $k=n$. The classifier in (\ref{formal}) with $\lambda\downarrow B_1/B_2$ (\textit{i.e.}, $\lambda_0\downarrow 0$) approaches the OWNN classifier considered in \citet{S12}. Note that the two terms $(n^{-2/d}\sum_{i=1}^n \alpha_i w_{ni})^2$ and $\sum_{i=1}^n w_{ni}^2$ in (\ref{formal}) represent the bias and variance terms of the regret expansion given in Proposition \ref{thm:regret} \citep{S12}. By varying the weights of these two terms through $\lambda$, we are able to stabilize a nearest neighbor classifier. Moreover, the stabilized classifier achieves desirable convergence rates in both regret and CIS; see Section~\ref{sec:thms}.

Theorem \ref{thm:optimal} gives the optimal weight $w_{ni}^*$ with respect to the optimization (\ref{formal}). We formally define the stabilized nearest neighbor (SNN) classifier as the WNN classifier with the optimal weight $w_{ni}^*$.
\begin{theorem}
\label{thm:optimal}
(Optimal Weight) For any fixed $\lambda>0$, the minimizer of $(\ref{formal})$ is
\[ w_{ni}^* = \left\{ \begin{array}{ll}
         \frac{1}{k^*}\left(1+\frac{d}{2}-\frac{d}{2(k^*)^{2/d}}\alpha_i\right), & \mbox{for $i=1,\ldots,k^*$},\\
        0, & \mbox{for $i = k^*+1, \ldots, n$},\end{array} \right. \]
where $\alpha_i=i^{1+\frac{2}{d}}-(i-1)^{1+\frac{2}{d}}$ and  $k^*= \lfloor \{\frac{d(d+4)}{2(d+2)}\}^{\frac{d}{d+4}}\lambda^{\frac{d}{d+4}}n^{\frac{4}{d+4}}\rfloor.$
\end{theorem}

The SNN classifier encompasses the OWNN classifier as a special case when $\lambda=B_1/B_2$. 

The computational complexity of our SNN classifier is comparable to that of existing nearest neighbor classifiers. If we preselect a value for $\lambda$, SNN requires no training at all. The testing time consists of two parts: an $O(n)$ complexity for the computation of $n$ distances, where $n$ is the size of training data; and an $O(n\log n)$ complexity for sorting $n$ distances. The $k$NN classifier, for example, shares the same computational complexity. In practice, $\lambda$ is not predetermined and we may treat it as a tuning parameter, whose optimal value is selected via cross validation. See Algorithm 1 in Section~\ref{sec:algorithm} for details. We will show in Section~\ref{sec:algorithm} that the complexity of tuning in SNN is also comparable to existing methods.

\section{Theoretical Properties}\label{sec:thms}

\subsection{A Sharp Rate of CIS}\label{sec:sharprate}
Motivated by \citet{AT07}, we establish a sharp convergence rate of CIS for a general plug-in classifier. A plug-in classification procedure $\Psi$ first estimates the regression function $\eta(x)$ by $\widehat{\eta}_n(x)$, and then plugs it into the Bayes rule, that is, $\widehat\phi_n(x) = 1+ \ind{\widehat{\eta}_n(x)< 1/2}$.

The following \textit{margin condition} \citep{T04} is assumed for deriving the upper bound of the convergence rate, while two additional conditions are required for showing the lower bound. A distribution function $P$ satisfies the \textit{margin condition} if there exist constants $C_0>0$ and $\alpha\ge 0$ such that for any $\epsilon>0$,
\begin{align}
\mathbb P_X(0<|\eta(X)-1/2|\le \epsilon)\le C_0 \epsilon^{\alpha}.\label{margin}
\end{align}
The parameter $\alpha$ characterizes the behavior of the regression function $\eta$ near $1/2$, and a larger $\alpha$ implies a lower noise level and hence an easier classification scenario.

The second condition is on the smoothness of $\eta(x)$. Specifically, we assume that $\eta$ belongs to a \textit{H{\"o}lder class of functions} $\Sigma(\gamma,L,\mathbb R^d)$ (for some fixed $L, \gamma>0$) containing the functions $g:\mathbb R^d\rightarrow \mathbb R$ that are $\lfloor \gamma\rfloor$ times continuously differentiable and satisfy, for any $x,x'\in \mathbb R^d$,
$|g(x') - g_x(x')| \le L\|x-x'\|^{\gamma},$
where $\lfloor \gamma \rfloor$ is the largest integer not greater than $\gamma$, $g_x$ is the Taylor polynomial series of degree $\lfloor \gamma\rfloor$ at $x$, and $\|\cdot\|$ is the Euclidean norm.

Our last condition assumes that the marginal distribution $\bar P$ satisfies the \textit{strong density assumption}, defined in Supplementary \ref{sec:strongdensity}.

We first derive the rate of convergence of CIS by assuming an exponential convergence rate of the corresponding regression function estimator.
\begin{theorem}
\label{thm:upperCIS}
(Upper Bound) Let $\widehat{\eta}_n$ be an estimator of the regression function $\eta$ and let ${\cal R}\subset \mathbb R^d$ be a compact set. Let $\mathcal P$ be a set of probability distributions supported on ${\cal R} \times \{1,2\}$ such that for some constants $C_1, C_2>0$, some positive sequence $a_n\rightarrow \infty$, and almost all $x$ with respect to $\bar{P}$,
\begin{align}
\sup_{P\in {\cal P}} \mathbb P_{\cal D}\Big(|\widehat{\eta}_n(x)-\eta(x)| \ge \delta\Big) \le C_1 \exp(-C_2 a_n \delta^2) \label{exponential}
\end{align}
holds for any $n> 1$ and $\delta>0$, where $\mathbb P_{\cal D}$ is the probability with respect to $P^{\otimes n}$. Furthermore, if all the distributions $P\in\cal P$ satisfy the margin condition for a constant $C_0$, then the plug-in classification procedure $\Psi$ corresponding to $\widehat{\eta}_n$ satisfies
$$
\sup_{P\in {\cal P}} \textrm{CIS}(\Psi) \le C a_n^{-\alpha/2},
$$
for any $n> 1$ and some constant $C>0$ depending only on $\alpha,C_0, C_1$, and $C_2$.
\end{theorem}
It is worth noting that the condition in $(\ref{exponential})$ holds for various types of estimators. For example, Theorem 3.2 in \citet{AT07} showed that the local polynomial estimator satisfies $(\ref{exponential})$ with $a_n=n^{2\gamma/(2\gamma+d)}$ when the bandwidth is of the order $n^{-1/(2\gamma+d)}$. In addition, Theorem \ref{thm:upperCISsnn} in Section \ref{sec:ratesnn} implies that $(\ref{exponential})$ holds for the newly proposed SNN classifier with the same $a_n$. Hence, in both cases, the upper bound is of the order $n^{-\alpha\gamma/(2\gamma+d)}$.

We next derive the lower bound of CIS in Theorem \ref{thm:lowerCIS}. As will be seen, this lower bound implies that the obtained rate of CIS, that is, $n^{-\alpha\gamma/(2\gamma+d)}$, cannot be further improved for the plug-in classification procedure.

\begin{theorem}
\label{thm:lowerCIS}
(Lower Bound) Let ${\cal P}_{\alpha,\gamma}$ be a set of probability distributions supported on ${\cal R} \times \{1,2\}$ such that for any $P\in{\cal P}_{\alpha,\gamma}$, $P$ satisfies the margin condition $(\ref{margin})$, the regression function $\eta(x)$ belongs to the H\"older class $\Sigma(\gamma,L,\mathbb R^d)$, and the marginal distribution $\bar{P}$ satisfies the strong density assumption. Suppose further that ${\cal P}_{\alpha,\gamma}$ satisfies $(\ref{exponential})$ with $a_n=n^{2\gamma/(2\gamma+d)}$ and $\alpha\gamma\le d$. We have
$$
\sup_{P\in {\cal P}_{\alpha,\gamma}} \textrm{CIS}(\Psi) \ge C' n^{-\alpha\gamma/(2\gamma+d)},
$$
for any $n> 1$ and some constant $C'>0$ independent of $n$.
\end{theorem}

Theorems \ref{thm:upperCIS} and \ref{thm:lowerCIS} together establish a sharp convergence rate of the CIS for the general plug-in classification procedure on the set ${\cal P}_{\alpha,\gamma}$. The requirement $\alpha\gamma\le d$ in Theorem \ref{thm:lowerCIS} implies that $\alpha$ and $\gamma$ cannot be large simultaneously. As pointed out in \citet{AT07}, this is intuitively true because a very large $\gamma$ implies a very smooth regression function $\eta$, while a large $\alpha$ implies that $\eta$ cannot stay very long near $1/2$, and hence when $\eta$ hits $1/2$, it should take off quickly. Lastly, we note that this rate is slower than $n^{-1}$, but approaches $n^{-1}$ as the dimension $d$ increases when $\alpha\gamma=d$.

As a reminder, \citet{AT07} established the minimax optimal rate of regret as $n^{-(\alpha+1)\gamma/(2\gamma+d)}$.

\subsection{Optimal Convergence Rates of SNN}
\label{sec:ratesnn}

In this subsection, we demonstrate that SNN attains the established sharp convergence rate in CIS in the previous subsection, as well as the minimax optimal convergence rate in regret. We further show the asymptotic difference between SNN and OWNN.

In Theorem \ref{thm:upperCISsnn} and Corollary \ref{cor:snn_ownn} below, we consider $\textrm{SNN}$ with $k^* \asymp n^{2\gamma/(2\gamma+d)}$ in Theorem \ref{thm:optimal}, where $a_n \asymp b_n$ means the ratio sequence $a_n/b_n$ stays away from zero and infinity as $n\rightarrow\infty$. Note that under Assumptions (A1)--(A4) defined in Appendix \ref{sec:assumptions}, we have $\gamma=2$ and hence $k^* \asymp n^{4/(4+d)}$, which agrees with the formulation in Theorem \ref{thm:optimal}.

\begin{theorem}
\label{thm:upperCISsnn}
For any $\alpha\ge 0$ and $\gamma\in (0,2]$, the SNN procedure with any fixed $\lambda>0$ satisfies
\begin{eqnarray*}
\sup_{P\in {\cal P}_{\alpha,\gamma}} \textrm{Regret}(\textrm{SNN}) &\le& \tilde{C} n^{-(\alpha+1)\gamma/(2\gamma+d)},\\
\sup_{P\in {\cal P}_{\alpha,\gamma}} \textrm{CIS}(\textrm{SNN}) &\le& C n^{-\alpha\gamma/(2\gamma+d)},
\end{eqnarray*}
for any $n> 1$ and some constants $\tilde{C}, C>0$, where ${\cal P}_{\alpha,\gamma}$ is defined in Theorem~\ref{thm:lowerCIS}.
\end{theorem}

Corollary \ref{cor:snn_ownn} below further investigates the difference between the SNN procedure (with $\lambda\ne B_1/B_2$)  and the OWNN procedure in terms of both regret and CIS.

\begin{corollary}
\label{cor:snn_ownn}
For any $\alpha\ge 0$, $\gamma\in (0,2]$, we have, when $\lambda\ne B_1/B_2$,
\begin{eqnarray}
\sup_{P\in {\cal P}_{\alpha,\gamma}} \Big\{\textrm{Regret}(\textrm{SNN})-\textrm{Regret}(\textrm{OWNN})\Big\} &\asymp& n^{-(1+\alpha)\gamma/(2\gamma+d)},\nonumber\\
\sup_{P\in {\cal P}_{\alpha,\gamma}} \Big\{ \textrm{CIS}(\textrm{OWNN}) - \textrm{CIS}(\textrm{SNN})\Big\} &\asymp& n^{-\alpha\gamma/(2\gamma+d)}, \label{eqn:cisdifference}
\end{eqnarray}
where ${\cal P}_{\alpha,\gamma}$ is defined in Theorem~\ref{thm:lowerCIS}.
\end{corollary}
Corollary \ref{cor:snn_ownn} implies that the regret of SNN approaches that of the OWNN (from above) at a faster rate than the CIS of OWNN approaches that of the SNN procedure (from above). This means that SNN can have a significant improvement in CIS over the OWNN procedure while obtaining a comparable classification accuracy. This observation will be supported by the experimental results in Section \ref{sec:simulation}.

\begin{remark}
Under Assumptions (A1)--(A4) in Section \ref{sec:assumptions}, which implicitly implies that $\alpha=1$, and the assumption that $\gamma=2$, the conclusion in $(\ref{eqn:cisdifference})$ can be strengthened to that for any $P\in {\cal P}_{1,2}$, $\textrm{CIS}(\textrm{OWNN}) - \textrm{CIS}(\textrm{SNN}) \asymp n^{-2/(d+4)}$. It indicates that SNN's improvement in CIS is at least $n^{-2/(d+4)}$ in this scenario.

\end{remark}

\section{Asymptotic Comparisons}\label{theoretical_compare}
In this section, we first conduct an asymptotic comparison of CIS among existing nearest neighbor classifiers, and then demonstrate that SNN significantly improves OWNN in CIS.
\subsection{CIS Comparison of Existing Methods}\label{sec:ciscom}
We compare $k$NN, OWNN and the bagged nearest neighbor (BNN) classifier. The $k$NN classifier is a special case of the WNN classifier with weight $w_{ni}=1/k$ for $i=1,\ldots,k$ and $w_{ni}=0$ otherwise. Another special case of the WNN classifier is the BNN classifier. After generating subsamples from the original data set, the BNN classifier applies 1-nearest neighbor classifier to each bootstrapped subsample and returns the final prediction by majority voting. If the resample size $m$ is sufficiently smaller than $n$, \textit{i.e.}, $m\rightarrow \infty$ and $m/n\rightarrow 0$, the BNN classifier is shown to be a consistent classifier \citep{HS05}. In particular, \citet{HS05} showed that, for large $n$, the BNN classifier (with or without replacement) is approximately equivalent to a WNN classifier with the weight $w_{ni} = q(1-q)^{i-1}/[1-(1-q)^n]$ for $i=1,\ldots,n$, where $q$ is the resampling ratio $m/n$.

We denote the CIS of the above classification procedures as $\textrm{CIS}(\textrm{$k$NN})$, $\textrm{CIS}(\textrm{BNN})$ and $\textrm{CIS}(\textrm{OWNN})$. Here $k$ in the $k$NN classifier is selected as the one minimizing the regret \citep{HPS08}. The optimal $q$ in the BNN classifier and the optimal weight in the OWNN classifier are both calculated based on their asymptotic relations with the optimal $k$ in $k$NN, which were defined in (2.9) and (3.5) of \citet{S12}. Corollary \ref{cor:cisratio} gives the pairwise CIS ratios of these classifiers. Note that these ratios depend on the feature dimension $d$ only.

\begin{corollary}
\label{cor:cisratio}
Under Assumptions (A1)-(A4) defined in Appendix \ref{sec:assumptions} and the assumption that $B_2$ defined in Appendix \ref{sec:defwnb} is positive, we have, as $n\rightarrow \infty$,
\begin{eqnarray*}
\frac{\textrm{CIS}(\textrm{OWNN})}{\textrm{CIS}(\textrm{$k$NN})} &\longrightarrow& 2^{2/(d+4)}\Big(\frac{d+2}{d+4}\Big)^{(d+2)/(d+4)},\\
\frac{\textrm{CIS}(\textrm{BNN})}{\textrm{CIS}(\textrm{$k$NN})} &\longrightarrow& 2^{-2/(d+4)}\Gamma(2+2/d)^{d/(d+4)}, \\
\frac{\textrm{CIS}(\textrm{BNN})}{\textrm{CIS}(\textrm{OWNN})} &\longrightarrow& 2^{-4/(d+4)}\Gamma(2+2/d)^{d/(d+4)}\Big(\frac{d+4}{d+2}\Big)^{(d+2)/(d+4)}.
\end{eqnarray*}
\end{corollary}

\begin{figure}[!b]
\begin{center}\vspace{-3em}
\includegraphics[scale=0.6]{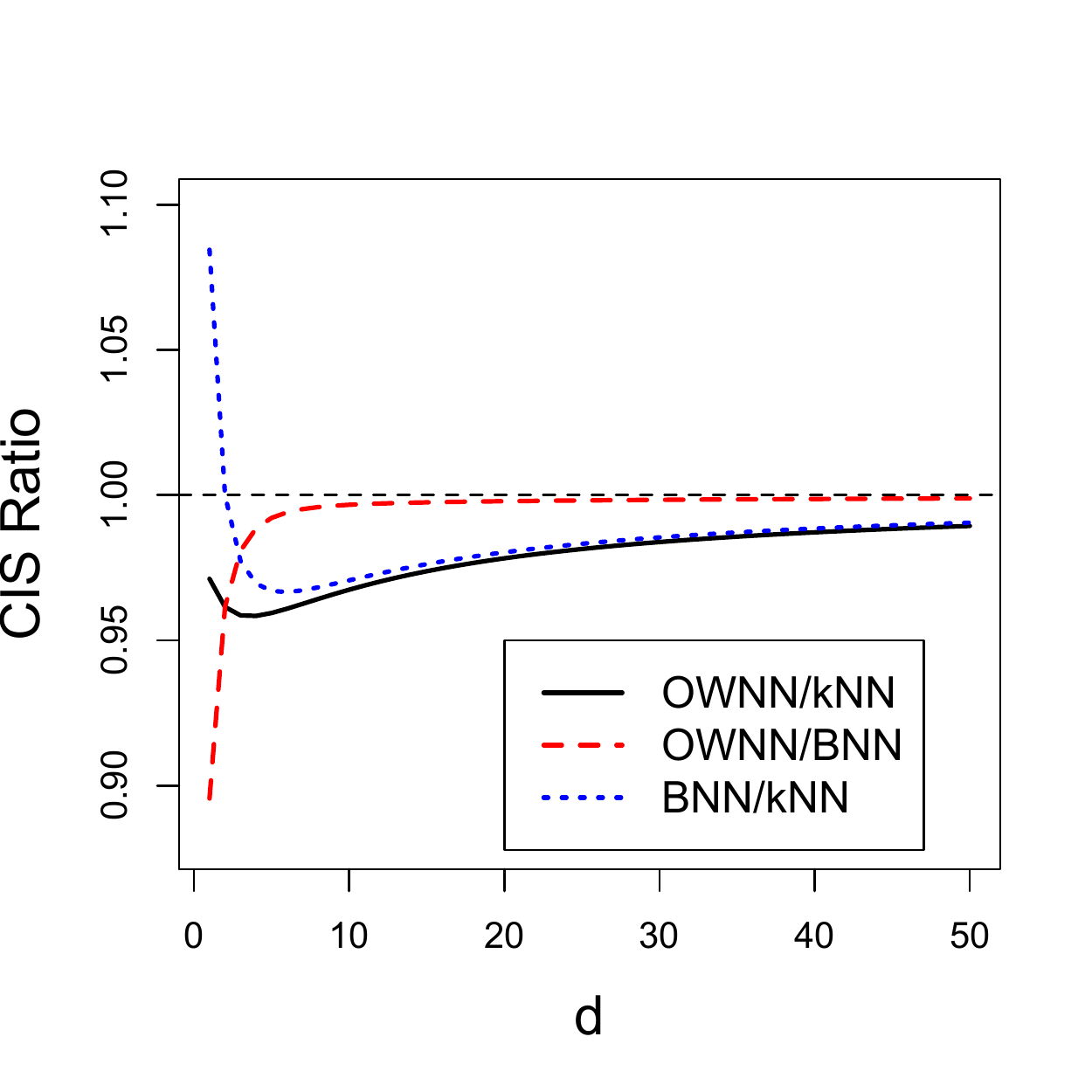}\vspace{-1em}
\caption{\label{CISratio_ownn_knn_bnn} \footnotesize Pairwise CIS ratios between $k$NN, BNN and OWNN for different feature dimension $d$.}
\end{center}
\end{figure}

The limiting CIS ratios in Corollary \ref{cor:cisratio} are plotted in Figure \ref{CISratio_ownn_knn_bnn}. A major message herein is that the OWNN procedure is more stable than the $k$NN and BNN procedures for any $d$. The largest improvement of the OWNN procedure over $k$NN is achieved when $d=4$ and the improvement diminishes as $d\rightarrow \infty$. The CIS ratio of BNN over $k$NN equals $1$ when $d=2$ and is less than $1$ when $d>2$, which is consistent with the common perception that bagging can generally reduce the variability of the nearest neighbor classifiers. Similar phenomenon has been shown in the ratio of their regrets \citep{S12}. Therefore, bagging can be used to improve the $k$NN procedure in terms of both accuracy and stability when $d>2$. Furthermore, the CIS ratio of OWNN over BNN is less than $1$ for all $d$, but quickly converges to $1$ as $d$ increases. This implies that although the BNN procedure is asymptotically less stable than the OWNN procedure, their difference vanishes as $d$ increases.

\subsection{Comparisons between SNN and OWNN}\label{sec:regcis}
Corollary~\ref{cor:snn_ownn} in Section \ref{sec:ratesnn} implies that OWNN and SNN have the same convergence rates of regret and CIS (note that OWNN is a special case of SNN). Hence, it is of more interest to compare their relative magnitude. The asymptotic comparisons between SNN and OWNN are characterized in Corollary \ref{cor:snn_ownn_ratio}.

\begin{corollary}
\label{cor:snn_ownn_ratio}
Under Assumptions (A1)-(A4) defined in Appendix \ref{sec:assumptions} and the assumption that $B_2$ defined in Appendix \ref{sec:defwnb} is positive, we have, as $n\rightarrow \infty$,
\begin{eqnarray*}
\frac{\textrm{Regret}(\textrm{SNN})}{\textrm{Regret}(\textrm{OWNN})} &\longrightarrow& \Big\{\frac{B_1}{\lambda B_2}\Big\}^{d/(d+4)}\Big\{\frac{4+d\lambda B_2/B_1}{4+d}\Big\},\\
\frac{\textrm{CIS}(\textrm{SNN})}{\textrm{CIS}(\textrm{OWNN})} &\longrightarrow& \Big\{\frac{B_1}{\lambda B_2}\Big\}^{d/(2(d+4))}, \nonumber
\end{eqnarray*}
where constants $B_1$ and $B_2$ are defined in Appendix \ref{sec:defwnb}.
\end{corollary}
The second formula in Corollary~\ref{cor:snn_ownn_ratio} suggests that as $\lambda$ increases, the SNN classifier becomes more and more stable. In Corollary~\ref{cor:snn_ownn_ratio}, both ratios of the SNN procedure over the OWNN procedure depend on $\lambda$, and two unknown constants $B_1$ and $B_2$. Since $\lambda = (B_1+\lambda_0B_3^2)/B_2$ in $(\ref{formal})$ and $B_3=4B_1/\sqrt{\pi}$ in $(\ref{eq:asy_CIS})$, we further have the following ratios,
\begin{eqnarray}
\frac{\textrm{Regret}(\textrm{SNN})}{\textrm{Regret}(\textrm{OWNN})} &\longrightarrow& \Big\{\frac{1}{1 + 16B_1\lambda_0/\pi}\Big\}^{d/(d+4)}\Big\{\frac{4+d(1+16B_1\lambda_0/\pi)}{4+d}\Big\}, \label{snn_ownn_regret}\\
\frac{\textrm{CIS}(\textrm{SNN})}{\textrm{CIS}(\textrm{OWNN})} &\longrightarrow& \Big\{\frac{1}{1 + 16B_1\lambda_0/\pi}\Big\}^{d/(2(d+4))}. \label{snn_ownn_cis}
\end{eqnarray}
For any $\lambda_0>0$, SNN has an improvement in CIS over the OWNN. As a mere illustration, we consider the case that the regret and the squared CIS are given equal weight, that is, $\lambda_0=1$. In this case, the ratios in $(\ref{snn_ownn_regret})$ and $(\ref{snn_ownn_cis})$ only depend on $B_1$ and $d$.

\begin{figure}[h!]
\begin{center}
\includegraphics[scale=0.55]{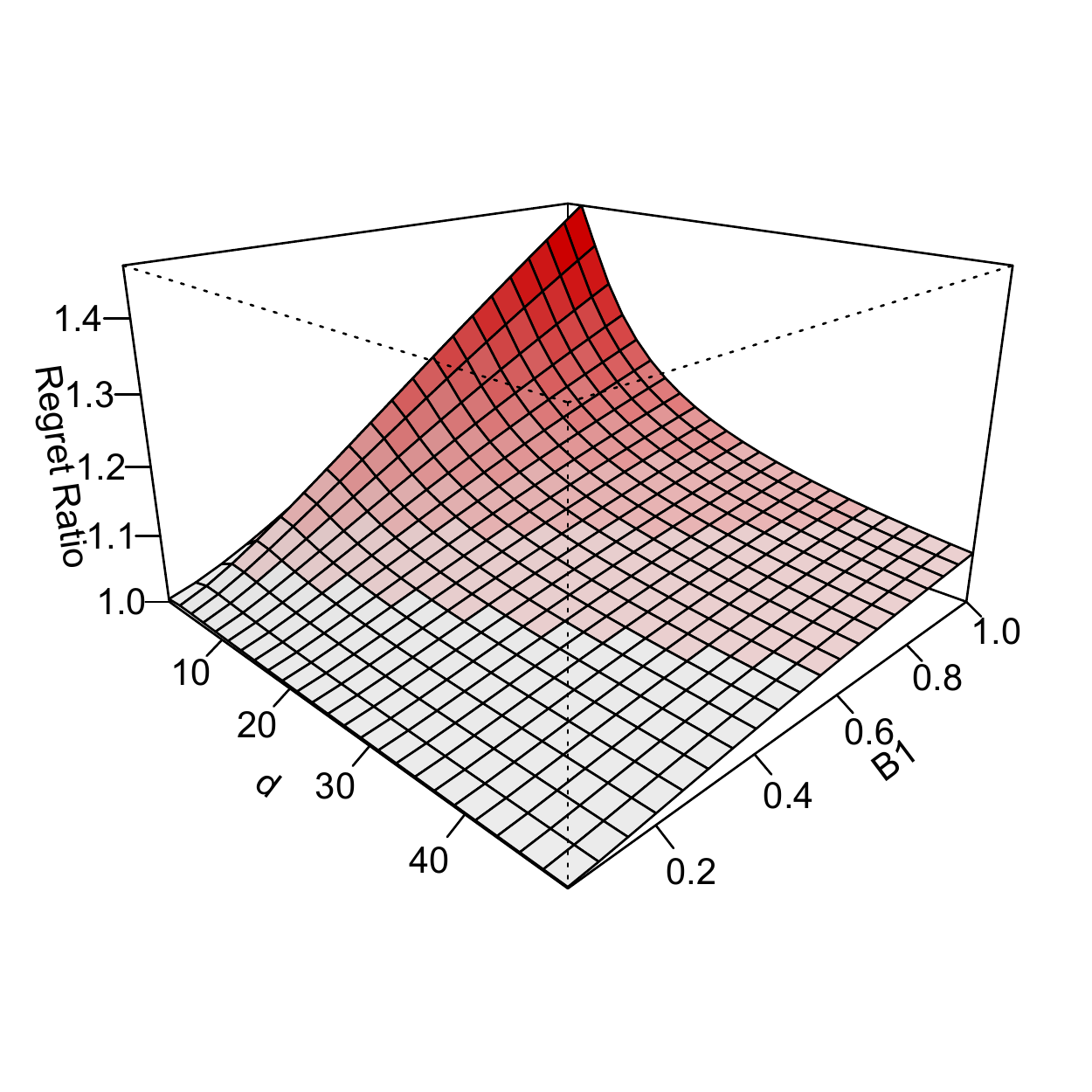}
\includegraphics[scale=0.55]{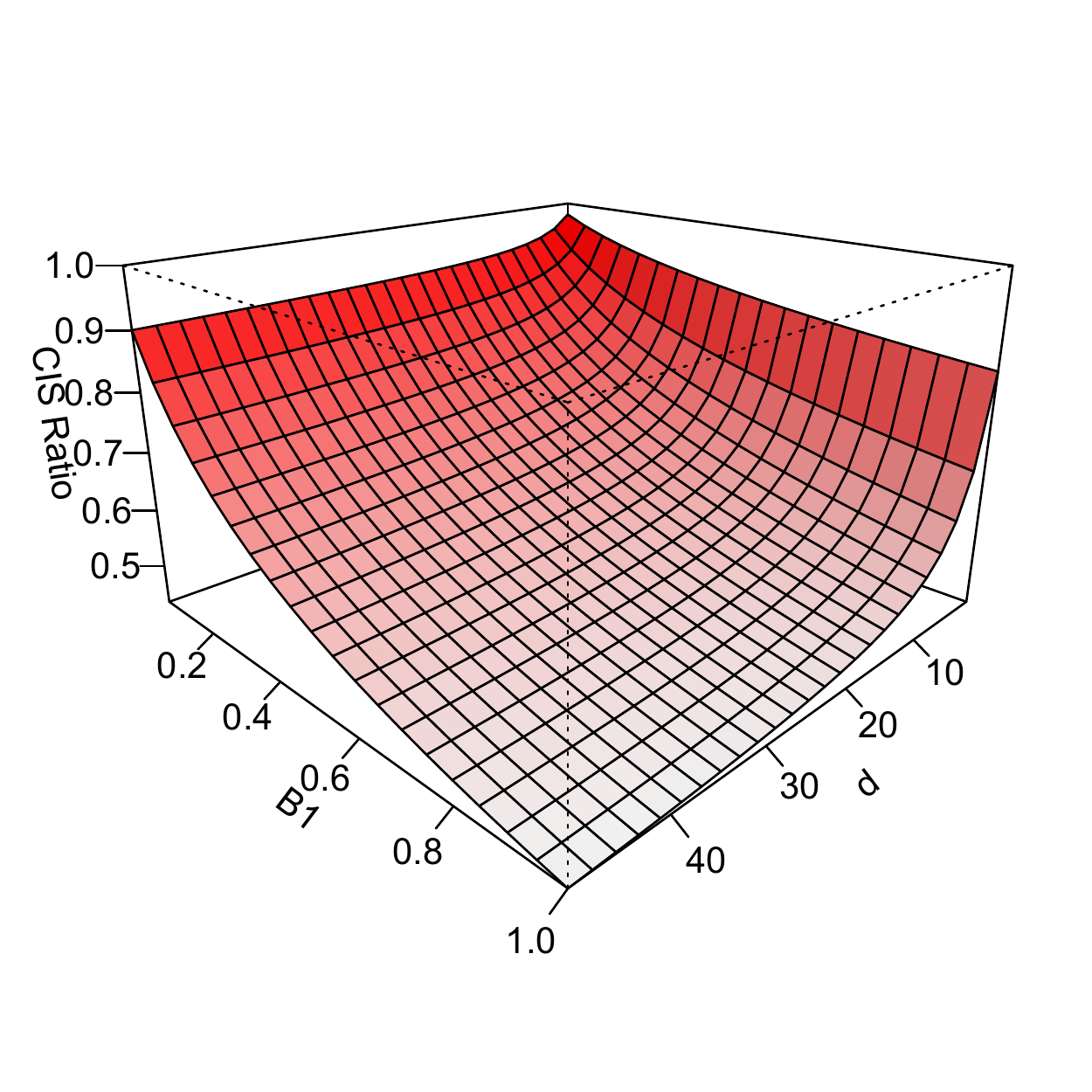}
\caption{\label{ratio_po} \footnotesize Regret ratio and CIS ratio of SNN over OWNN as functions of $B_1$ and $d$. The darker the color, the larger the value. }
\end{center}
\end{figure}

Figure \ref{ratio_po} shows 3D plots of these two ratios as functions of $B_1$ and $d$. As expected, the CIS of the SNN procedure is universally smaller than OWNN (ratios less than 1 on the right panel), while the OWNN procedure has a smaller regret (ratios greater than 1 on the left panel). For a fixed $B_1$, as the dimension $d$ increases, the regret of SNN approaches that of OWNN, while the advantage of SNN in terms of CIS grows. For a fixed dimension $d$, as $B_1$ increases, the regret ratio between SNN and OWNN gets larger, but the CIS advantage of SNN also grows. According to the definition of $B_1$ in Appendix \ref{sec:defwnb}, a great value of $B_1$ indicates a harder problem for classification; see the discussion after Theorem 1 of \cite{S12}.
\begin{figure}[b!]
\begin{center}\vspace{-2em}
\includegraphics[scale=0.55]{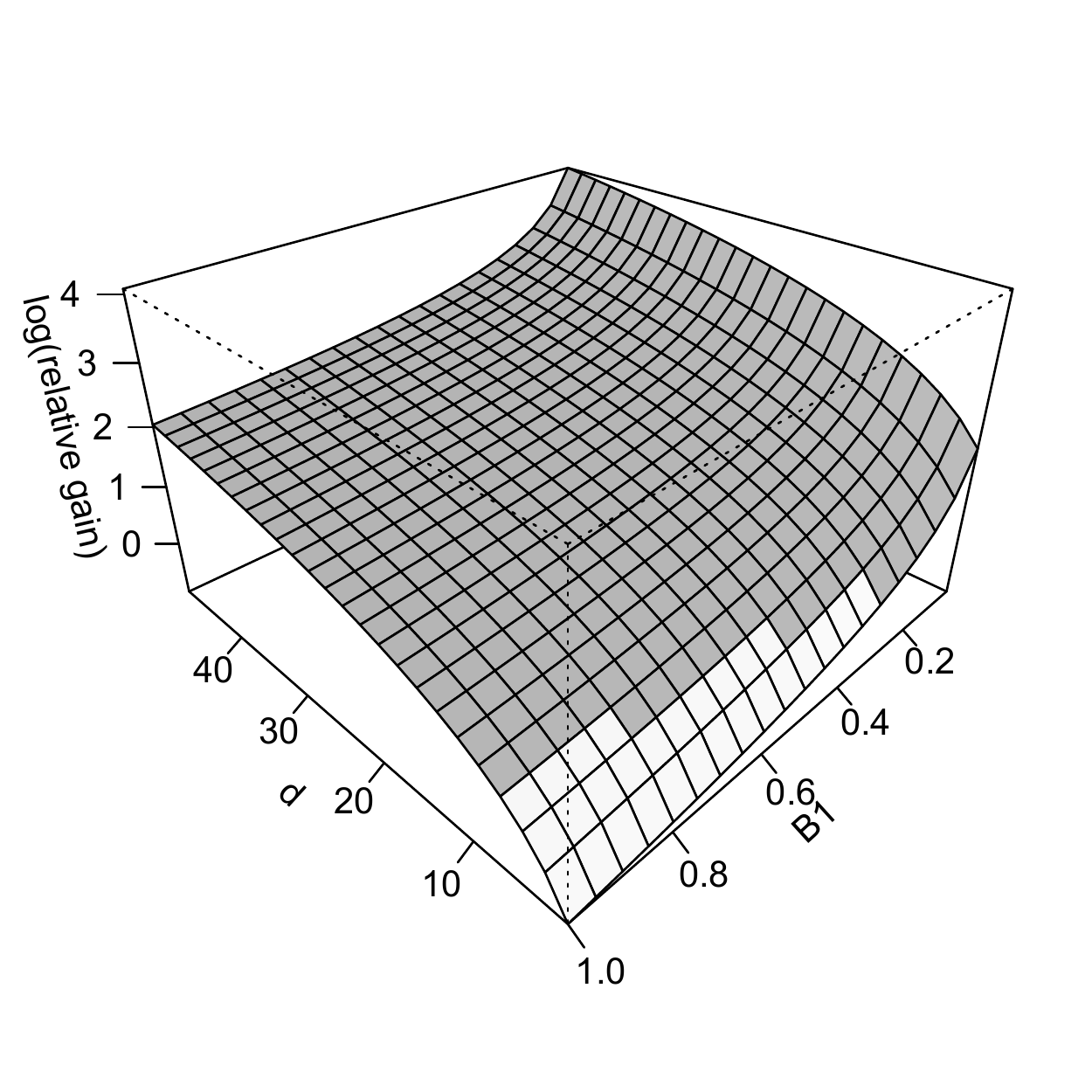}\vspace{-2em}
\caption{\footnotesize Logarithm of relative gain of SNN over OWNN as a function of $B_1$ and $d$ when $\lambda_0=1$. The grey (white) color represents the case where the logarithm of relative gain is greater (less) than $0$.}\label{loggain}
\end{center}
\end{figure}

Since SNN improves OWNN in CIS, but has a greater regret, it is of interest to know when the improvement of SNN in CIS is greater than its loss in regret. We thus consider the \textit{relative gain}, defined as the absolute ratio of the percentages of CIS reduction and regret increment, that is, $|\Delta \textrm{CIS}/\Delta \textrm{Regret}|$, where $\Delta \textrm{CIS} = [\textrm{CIS}(\textrm{SNN})-\textrm{CIS}(\textrm{OWNN})]/\textrm{CIS}(\textrm{OWNN})$ and $\Delta \textrm{Regret} = [\textrm{Regret}(\textrm{SNN})-\textrm{Regret}(\textrm{OWNN})]/\textrm{Regret}(\textrm{OWNN})$. As an illustration, when $\lambda_0=1$, we have the relative gain converges to $\left[1- (1+16B_1/\pi)^{-d/(2d+8)} \right]\left[ (1+16B_1/\pi)^{4/(d+4)} -1\right]^{-1}$. Figure \ref{loggain} shows the log(relative gain) as a function of $B_1$ and $d$. For most combinations of $B_1$ and $d$, the logarithm is greater than $0$ (shown in grey in Figure \ref{loggain}), indicating that SNN has an improvement in CIS greater than its loss in regret. In particular, when $B_1\le 0.2$, the log(relative gain) is positive for all $d$.

\section{Tuning Parameter Selection}\label{sec:algorithm}
To select the parameter $\lambda$ for the SNN classifier, we first identify a set of values for $\lambda$ whose corresponding (estimated) risks are among the smallest, and then choose from them an optimal one which has the minimal estimated CIS. Let $\widehat \phi^\lambda_{\mathcal D}$ denote an SNN classifier with parameter $\lambda$ trained from sample $\mathcal D$. Given a predetermined set of tuning parameter values $\Lambda = \{\lambda_1,\ldots,\lambda_K\}$, the tuning parameter $\widehat{\lambda}$ is selected using Algorithm 1 below, which involves estimating the CIS and risk in Steps 1--3 and a two-stage selection in Steps 4 and 5. \\
\indent {\it Algorithm 1:}\\
\indent {\it Step 1}. Randomly partition ${\cal D}=\{(X_i,Y_i),i=1,\ldots,n\}$ into five subsets $I_i$, $i=1,\cdots,5$.\\
\indent {\it Step 2}. For $i=1$, let $I_1$ be the test set and $I_2$, $I_3$, $I_4$ and $I_5$ be training sets. Obtain predicted labels from $\widehat{\phi}_{I_2\cup I_3}^{\lambda}(X_j)$ and $\widehat{\phi}_{I_4\cup I_5}^{\lambda}(X_j)$ respectively for each $X_j\in I_1$. Estimate the CIS and risk of the classifier with parameter $\lambda$ by
\begin{eqnarray*}
\widehat{\textrm{CIS}}_{i}(\lambda) &=& \frac{1}{|I_1|} \sum_{(X_j,Y_j)\in I_1} \ind{\widehat{\phi}_{I_2 \cup I_3}^{\lambda}(X_j) \ne \widehat{\phi}_{I_4\cup I_5}^{\lambda}(X_j)},\\
\widehat{\textrm{Risk}}_{i}(\lambda) &=& \frac{1}{2|I_1|} \sum_{(X_j,Y_j)\in I_1} \left\{ \ind{\widehat{\phi}_{I_2 \cup I_3}^{\lambda}(X_j) \ne Y_j} + \ind{\widehat{\phi}_{I_4\cup I_5}^{\lambda}(X_j)\ne Y_j}\right\}.
\end{eqnarray*}
\indent {\it Step 3}. Repeat {\it Step 2} for $i=2,\dots,5$ and estimate the CIS and risk, with $I_i$ being the test set and the rest being the training sets. Finally, the estimated CIS and risk are,
\begin{eqnarray*}
\widehat{\textrm{CIS}}(\lambda) = \frac{1}{5}\sum_{i=1}^{5} \widehat{\textrm{CIS}}_{i}(\lambda),~~
\widehat{\textrm{Risk}}(\lambda) = \frac{1}{5}\sum_{i=1}^{5} \widehat{\textrm{Risk}}_{i}(\lambda).
\end{eqnarray*}
\indent {\it Step 4}. Perform {\it Step 2} and {\it Step 3} for each $\lambda_k \in \Lambda$. Denote the set of tuning parameters with top accuracy as
$$
{\cal A} := \{ \lambda:   \widehat{\textrm{Risk}}(\lambda) \textrm{~is less than the 10th percentile of~} \widehat{\textrm{Risk}}(\lambda_k),~k=1,\ldots,K\}.
$$
\indent {\it Step 5}. Output the optimal tuning parameter $\widehat{\lambda}$ as
$$
\widehat{\lambda} = \argmin_{\lambda \in {\cal A}} \widehat{\textrm{CIS}}(\lambda).
$$

In our experiments, the predetermined set of tuning parameters $\Lambda$ is of size $100$. In Step 1, the sample sizes of the subsets $I_i$ are chosen to be roughly equal. In Step 4, the threshold $10\%$ reflects how the set of the most accurate classifiers is defined. Based on our limited experiments, the final experimental result is very robust to the choice of this threshold level within a suitable range.

Compared with the tuning method for the $k$NN classifier, which minimizes the estimated risk only, Algorithm 1 requires additional estimation of the CIS. However, the estimation of the CIS is concurrently conducted with the estimation of the risk in Step 2. Therefore, the complexity of tuning for our SNN classifier is at the same order as that for $k$NN. As will be seen in the numerical experiments below, the additional effort on estimating the CIS leads to improvement over existing nearest neighbor methods in both accuracy and stability.

\section{Numerical Studies}\label{sec:exp}

We first verify our theoretical findings using an example, and then illustrate the improvements of the SNN classifier over existing nearest neighbor classifiers based on simulations and real examples.

\subsection{Validation of Asymptotically Equivalent Forms}\label{sec:validation}

This subsection aims to support the asymptotically equivalent forms of CIS derived in Theorem \ref{thm:CIS} and the CIS and regret ratios in Corollary \ref{cor:snn_ownn_ratio}. We focus on a multivariate Gaussian example in which regret and CIS have explicit expressions.

Assume that the underlying distributions of both classes are $P_1\sim N(0_2,\mathbb{I}_2)$ and $P_2\sim N(1_2,\mathbb{I}_2)$ and the prior class probability $\pi_1=1/3$. We choose ${\cal R}=[-2,3]^2$, which covers at least $95\%$ probability of the sampling region, and set $n=50, 100, 200$ and $500$. In addition, a test set with 1000 observations was independently generated. The estimated risk and CIS were calculated based on $100$ replications. In this example, some calculus exercises lead to $B_1=0.1299$, $B_2=10.68$ and $B_3=0.2931$. According to Proposition \ref{thm:regret}, Theorems \ref{thm:CIS} and \ref{thm:optimal}, we obtain that
\begin{align}
\textrm{Regret}(\textrm{SNN}) &= 0.1732{(k^*)}^{-1} + 4.7467 (k^*)^2n^{-2} \label{regret_sce1}\\
\textrm{CIS}(\textrm{SNN}) &= 0.3385(k^*)^{-1/2}, \label{CIS_sce2}
\end{align}
with $k^*= \lfloor 1.5^{1/3} \lambda^{1/3} n^{2/3}\rfloor$. For a mere illustration, we choose $\lambda=(B_1+B_3^2)/B_2$, which corresponds to $\lambda_0=1$. So we have $k^*=\lfloor 0.3118 n^{2/3}\rfloor.$ 

Similarly, the asymptotic regret and CIS of OWNN are (\ref{regret_sce1}) and (\ref{CIS_sce2}) with $k^*= \lfloor 0.2633 n^{2/3}\rfloor$ due to $(2.4)$ in \citet{S12}.

\begin{figure}[htb]
\begin{center}
\includegraphics[scale=0.6]{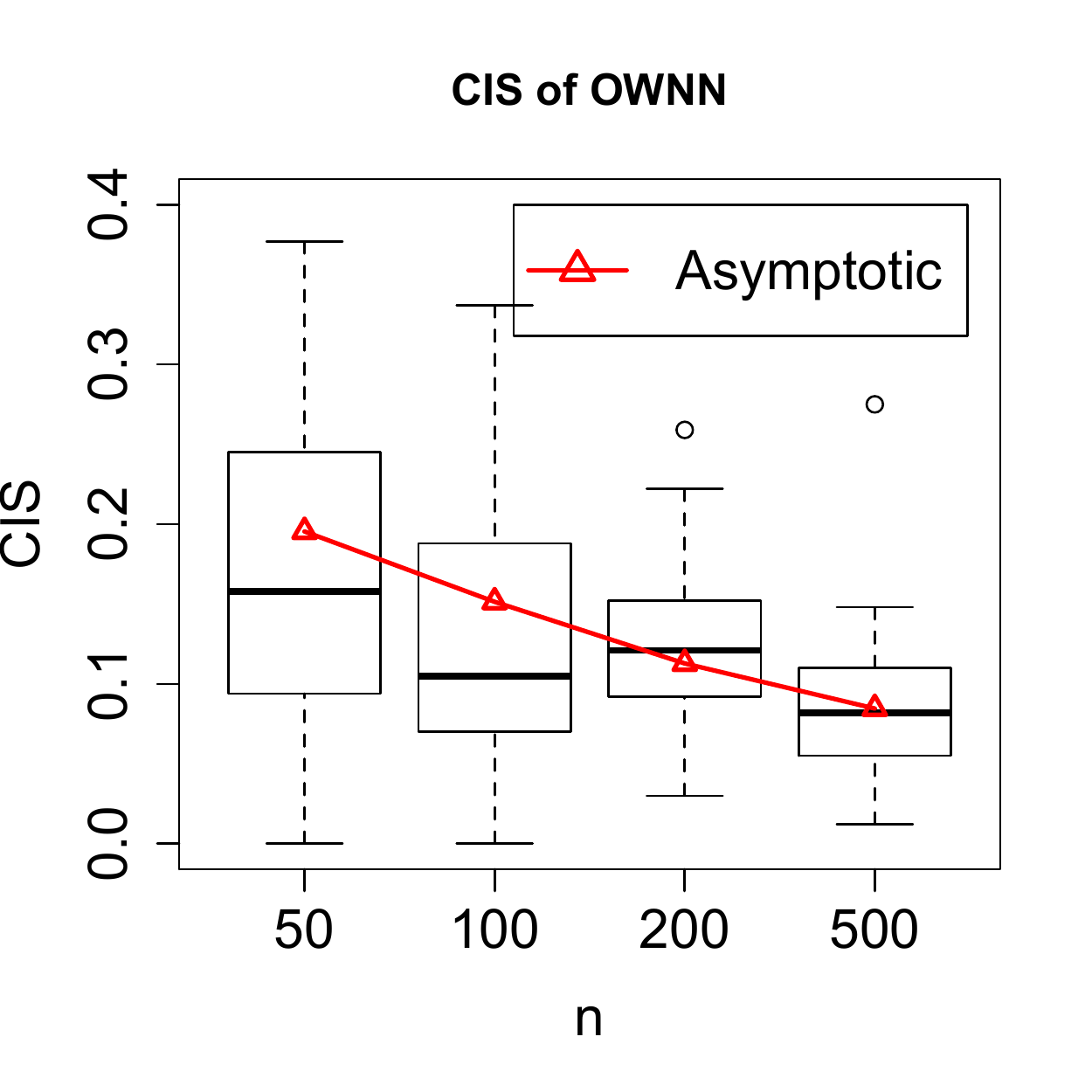}
\includegraphics[scale=0.6]{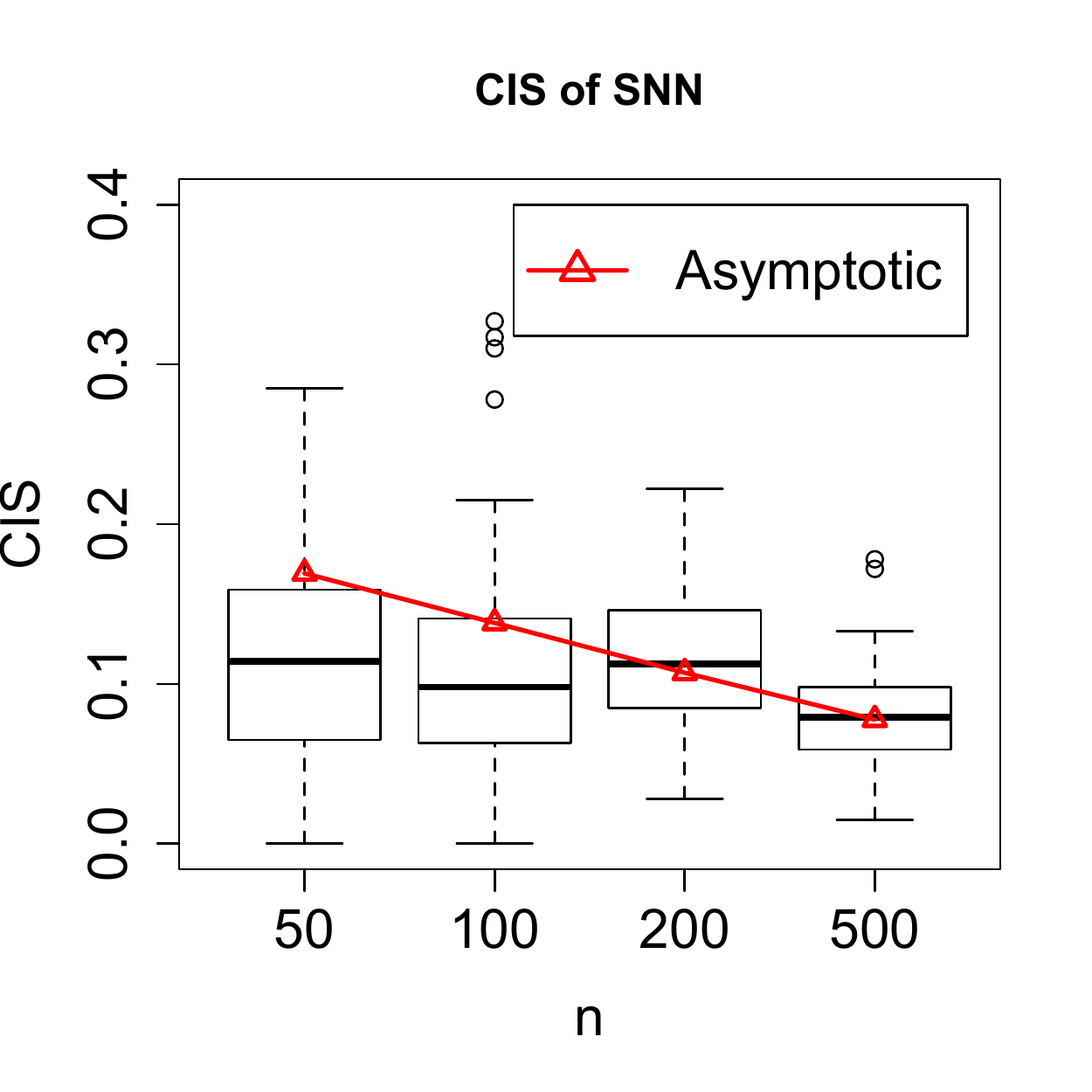}
\caption{\label{fig:CIS_sce1} \footnotesize Asymptotic CIS (red curve) and estimated CIS (box plots over 100 simulations) for OWNN (left) and SNN (right) procedures. These plots show that the estimated CIS converges to its asymptotic equivalent value as $n$ increases.}
\end{center}
\end{figure}

In Figures \ref{fig:CIS_sce1}, we plot the asymptotic CIS of the SNN and OWNN classifiers computed using the above formulae, shown as red curves, along with the estimated CIS based on the simulated data, shown as the box plots over 100 replications. As the sample size $n$ increases, the estimated CIS approximates its asymptotic value very well. For example, when $n=500$, the asymptotic CIS of the SNN (OWNN) classifier is 0.078 (0.085) while the estimated CIS is 0.079 (0.086).

\begin{figure}[!b]
\begin{center}
\includegraphics[scale=0.6]{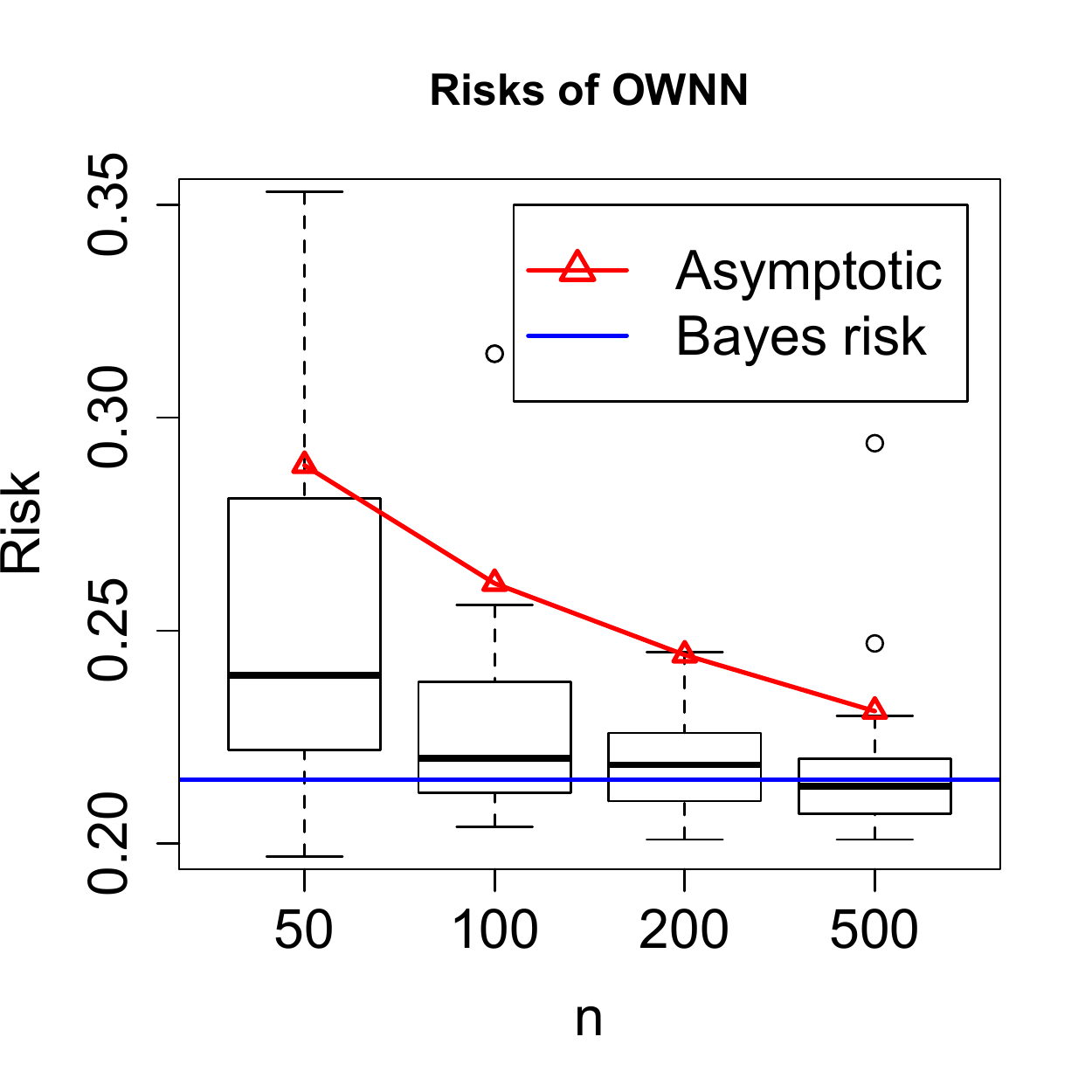}
\includegraphics[scale=0.6]{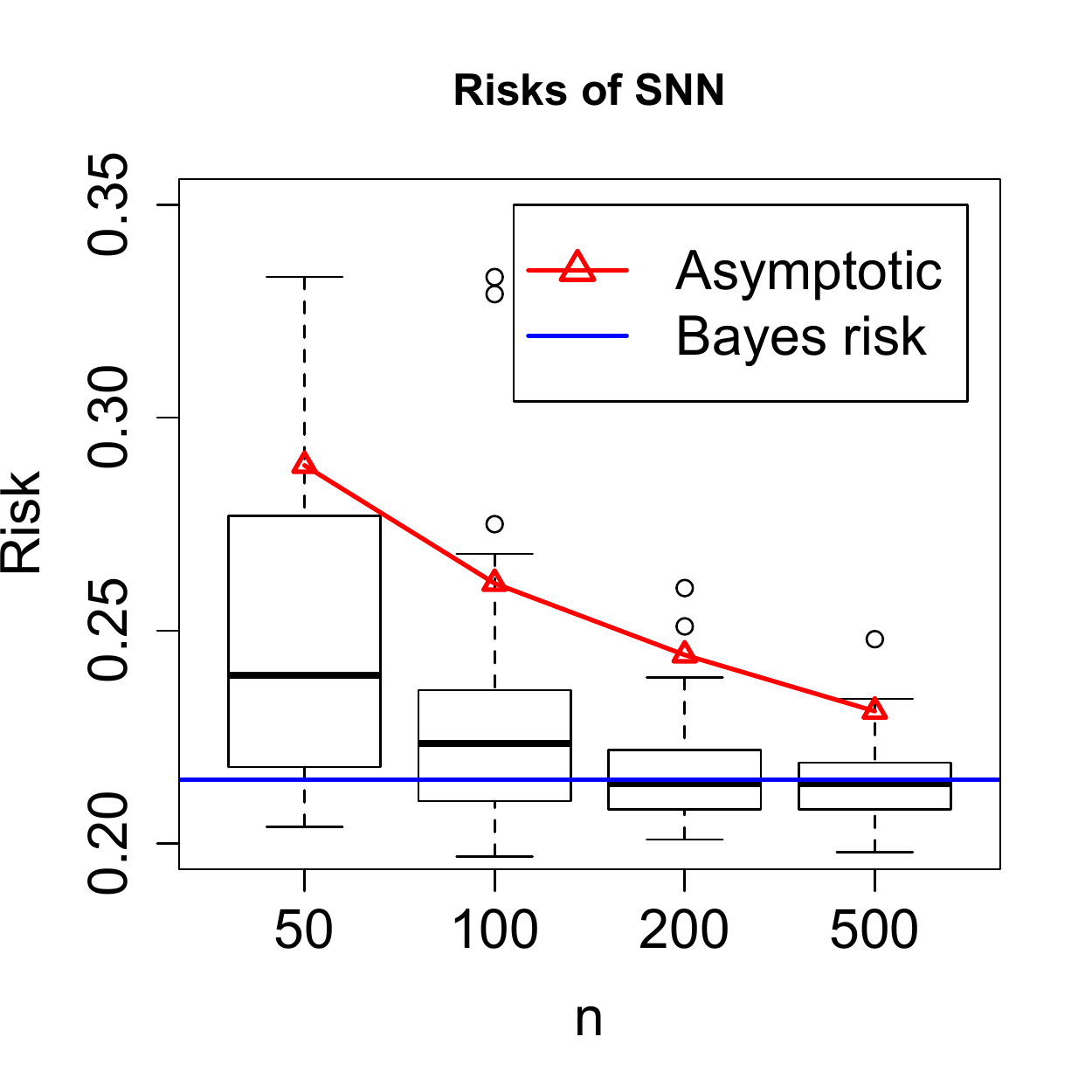}\vspace{-1em}
\caption{\label{fig:error_sce1} \footnotesize Asymptotic risk (regret + the Bayes risk; red curves) and estimated risk (black box plots) for OWNN (left) and SNN procedures (right). The blue horizontal line indicates the Bayes risk, 0.215. These plots show that the estimated risk converges to its asymptotic version (and also the Bayes risk) as $n$ increases.}
\end{center}
\end{figure}

Similarly, in Figure \ref{fig:error_sce1}, we plot the asymptotic risk, that is, the asymptotic regret in (\ref{regret_sce1}) plus the true Bayes risk ($0.215$ in this example), for the SNN and OWNN classifiers, along with the estimated risk. Here we compute the Bayes risk by Monte Carlo integration. Again the difference of the estimated risk and asymptotic risk decreases as the sample size grows.

Furthermore, according to $(\ref{snn_ownn_cis})$, the asymptotic CIS ratio of the SNN classifier over the OWNN classifier is $0.9189$ in this example, and the empirically estimated CIS ratios are $0.6646$, $0.9114$, $0.8940$ and $0.9219$, for $n=50,100,200,500$. This indicates that the estimated CIS ratio converges to its asymptotic value as $n$ increases. However, by $(\ref{snn_ownn_regret})$, the asymptotic regret ratio of the SNN classifier over the OWNN classifier is $1.0305$, while the estimated ones are $1.0224$, $1.1493$, $0.3097$ and $0.1136$,  for $n=50,100,200,500$. It appears that the estimated regret ratio matches with its asymptotic value for small sample size, but they differ for large $n$. This may be caused by the fact that the classification errors are very close to Bayes risk for large $n$ and hence the estimated regret ratio has a numerical issue. For example, when $n=500$, the average errors of the SNN classifier and the OWNN classifier are $0.2152$ and $0.2161$, respectively, while the Bayes risk is $0.215$ (see Figure \ref{fig:error_sce1}). A similar issue was previously reported in \citet{S12}.

\subsection{Simulations}
\label{sec:simulation}
In this section, we compare SNN with the $k$NN, OWNN and BNN classifiers. The parameter $k$ in $k$NN was tuned from $100$ equally spaced grid points from 5 to $n/2$. For a fair comparison, in the SNN classifier, the parameter $\lambda$ was tuned so that the corresponding parameter $k^*$ (see Theorem~\ref{thm:optimal}) were equally spaced and fell  into the same range roughly.

In Simulation 1, we assumed that the two classes were from $P_1\sim N(0_d,\mathbb{I}_d)$ and $P_2\sim N(\mu_d,\mathbb{I}_d)$ with the prior probability $\pi_1=1/3$ and dimension $d$. We set sample size $n=200$ and chose $\mu$ such that the resulting $B_1$ was fixed as $0.1$ for different $d$. Specifically, in Supplementary \ref{proof53} we show that
\begin{equation}
B_1 = \frac{\sqrt{2\pi}}{3\pi\mu d}\exp\left(-\frac{( \mu d/2-\ln 2/\mu)^2}{2d}\right). \label{B1}
\end{equation}
Hence, we set $\mu=2.076$, $1.205$, $0.659$, $0.314$, $0.208$ for $d=1,2,4,8$ and $10$, respectively.

In Simulation 2, the training data set were generated by setting $n=200$, $d=2$ or $5$, $P_1\sim 0.5N(0_d,\mathbb{I}_d)+0.5 N(3_d,2\mathbb{I}_d)$, $P_2\sim 0.5N(1.5_d,\mathbb{I}_d)+0.5 N(4.5_d,2\mathbb{I}_d)$, and $\pi_1=1/2$ or $1/3$.

Simulation 3 has the same setting as Simulation 2, except that $P_1\sim 0.5N(0_d,\Sigma)+0.5 N(3_d,2\Sigma)$ and $P_2\sim 0.5N(1.5_d,\Sigma)+0.5 N(4.5_d,2\Sigma)$, where $\Sigma$ is the Toeplitz matrix whose $j$th entry of the first row is $0.6^{j-1}$.

\begin{figure}[!t]
\begin{center}
\includegraphics[scale=0.5]{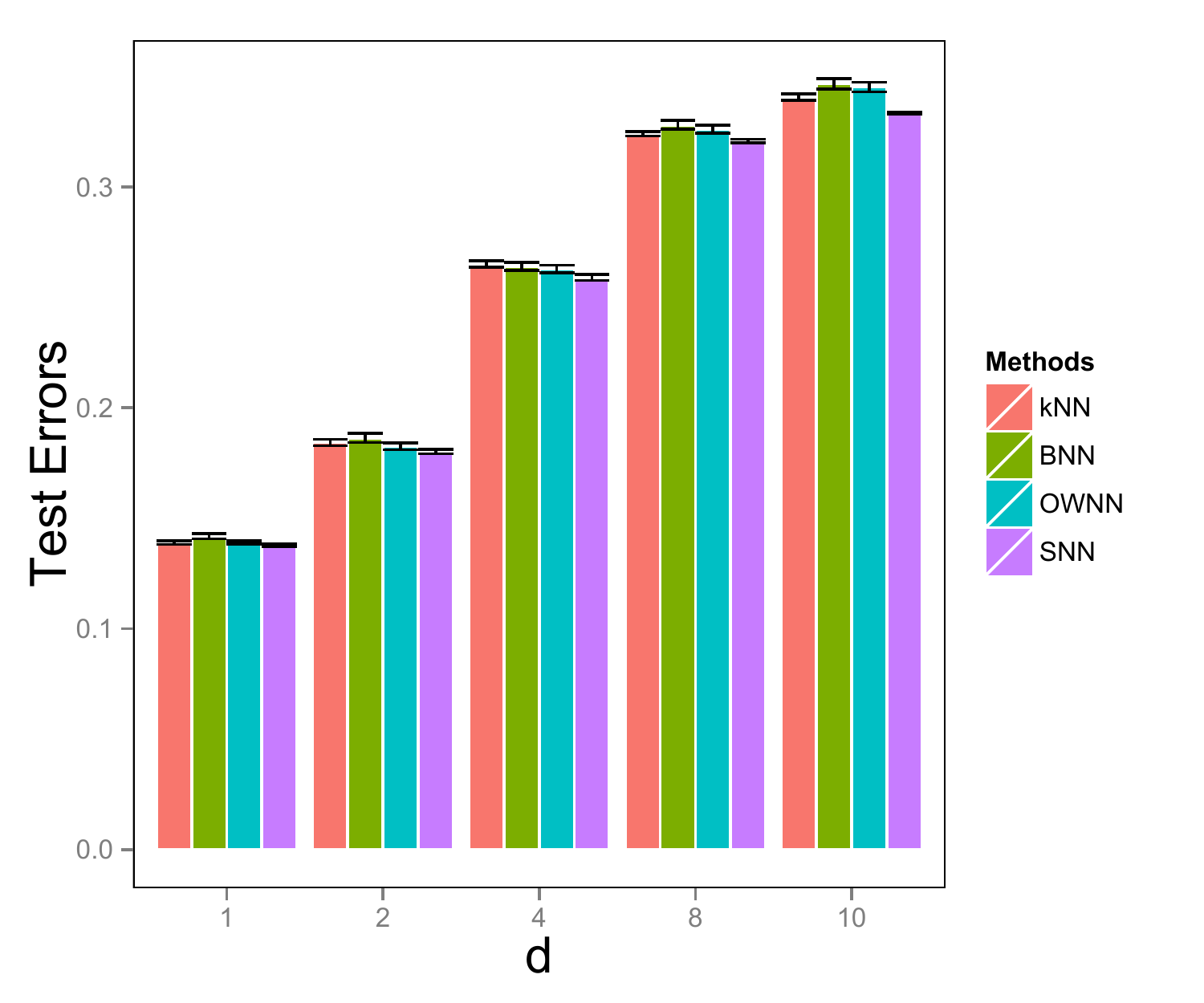}
\includegraphics[scale=0.5]{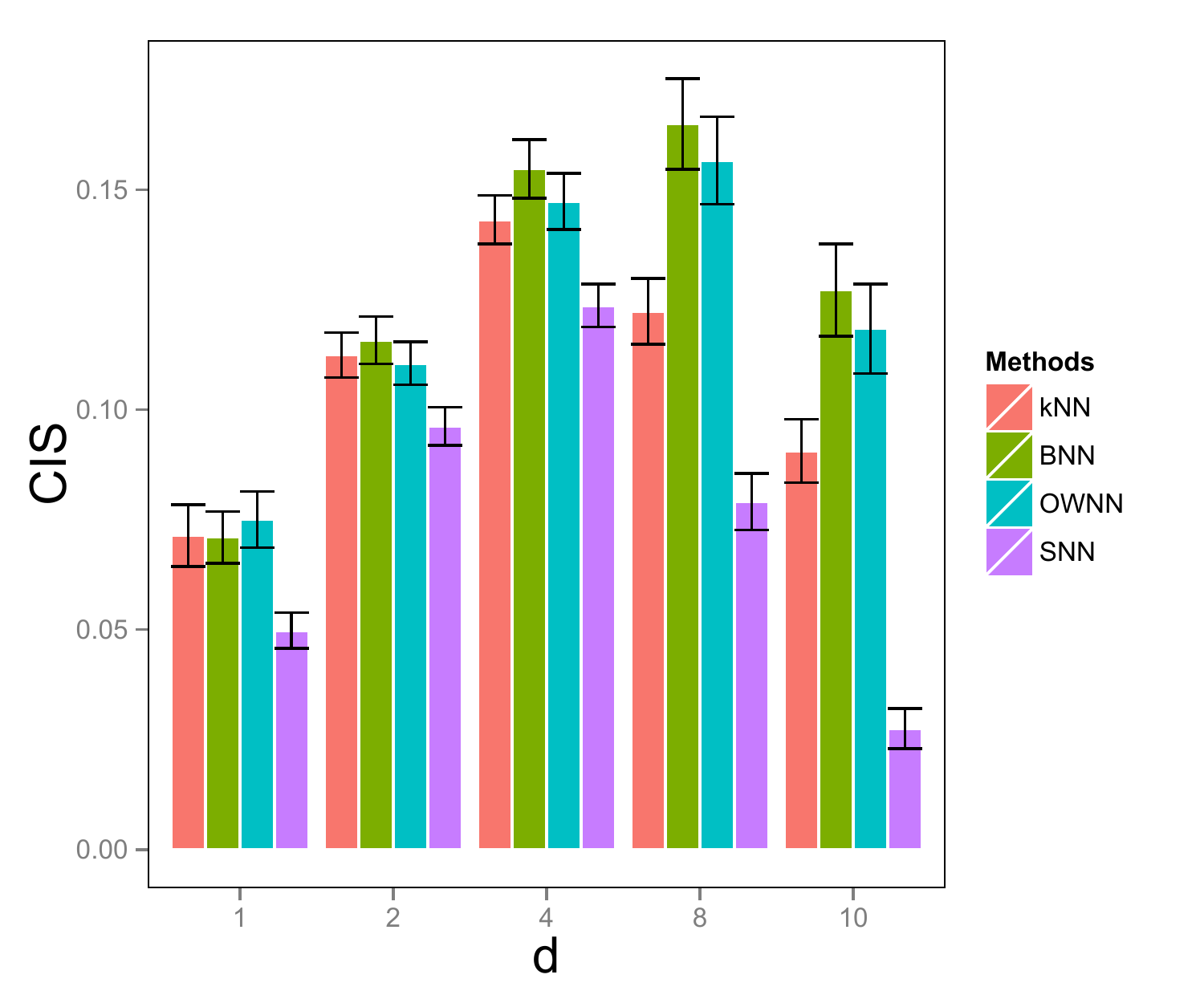}
\caption{\label{figure_sim1} \footnotesize Average test errors and CIS's (with standard error bar marked) of the $k$NN, BNN, OWNN, and SNN methods in Simulation 1. The $x$-axis indicates different settings with various dimensions. Within each setting, the four methods are horizontally lined up (from the left are $k$NN, BNN, OWNN, and SNN).}
\end{center}
\end{figure}

Simulation 1 is a relatively easy classification problem. Simulation 2 examines the bimodal effect and Simulation 3 combines bimodality with dependence between variables. In each simulation setting, a test data set of size $1000$ is independently generated and the average classification error and average estimated CIS for the test set are reported over $100$ replications. To estimate the CIS, for each replication, we build two classifiers based on the randomly divided training data, and then estimate CIS by the average disagreement of these two classifiers on the test data.

Figure \ref{figure_sim1} shows the average error (on the left) and CIS (on the right) for Simulation 1. As a first impression, the test error is similar among different classification methods, while the CIS differs a lot. In terms of the stability, SNN always has the smallest CIS; in particular, as $d$ increases, the improvement of SNN over all other procedures becomes even larger. This agrees with the asymptotic findings in Section \ref{sec:regcis}. For example, when $d=10$, all the $k$NN, BNN, and OWNN procedures are at least five times more unstable than SNN. In terms of accuracy, SNN obtains the minimal test errors in all five scenarios, although the improvement in the accuracy is not significant when $d=1,~2$ or $4$. This result suggests that although SNN is asymptotically less accurate than OWNN in theory, the actual empirical difference in the test error is often ignorable.

\begin{figure}[!t]
\begin{center}
\includegraphics[scale=0.5]{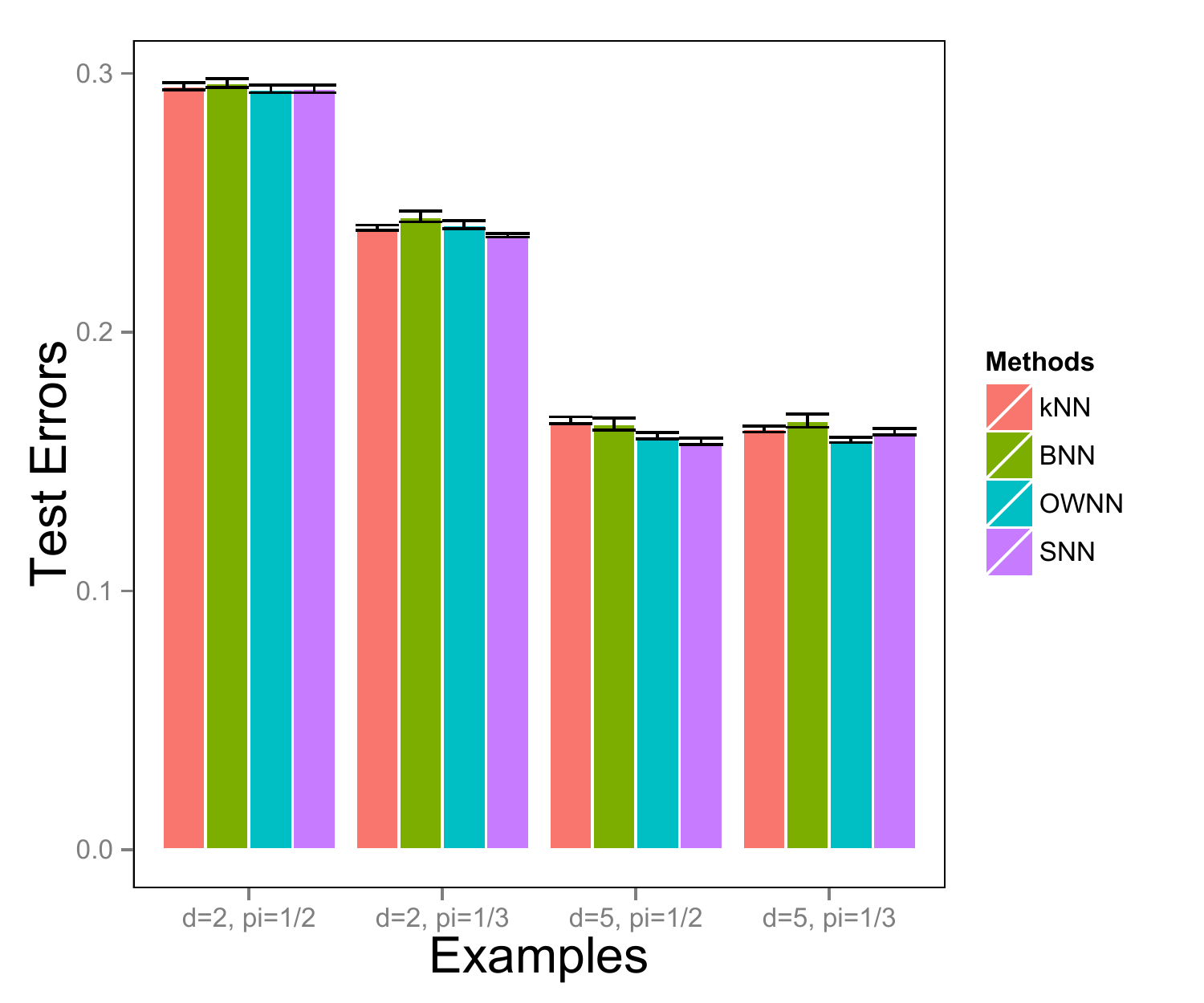}
\includegraphics[scale=0.5]{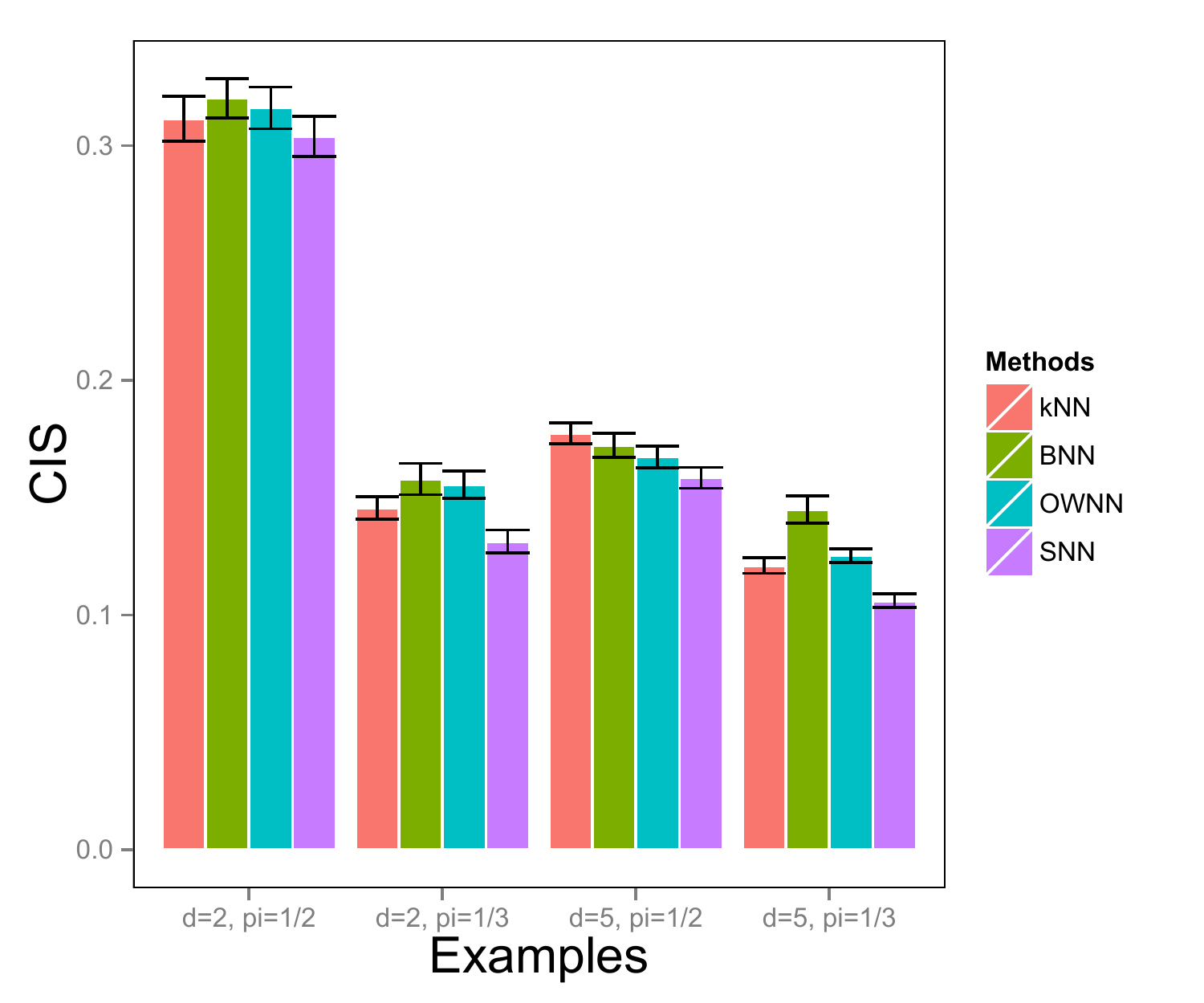}
\caption{\label{figure_sim2} \footnotesize Average test errors and CIS's  (with standard error bar marked) of the $k$NN, BNN, OWNN, and SNN methods in Simulation 2. The ticks on the $x$-axis indicate the dimensions and prior class probability $\pi$ for different settings. Within each setting, the four methods are horizontally lined up (from the left are $k$NN, BNN, OWNN, and SNN).}
\end{center}
\end{figure}

\begin{figure}[t!]
\begin{center}
\includegraphics[scale=0.5]{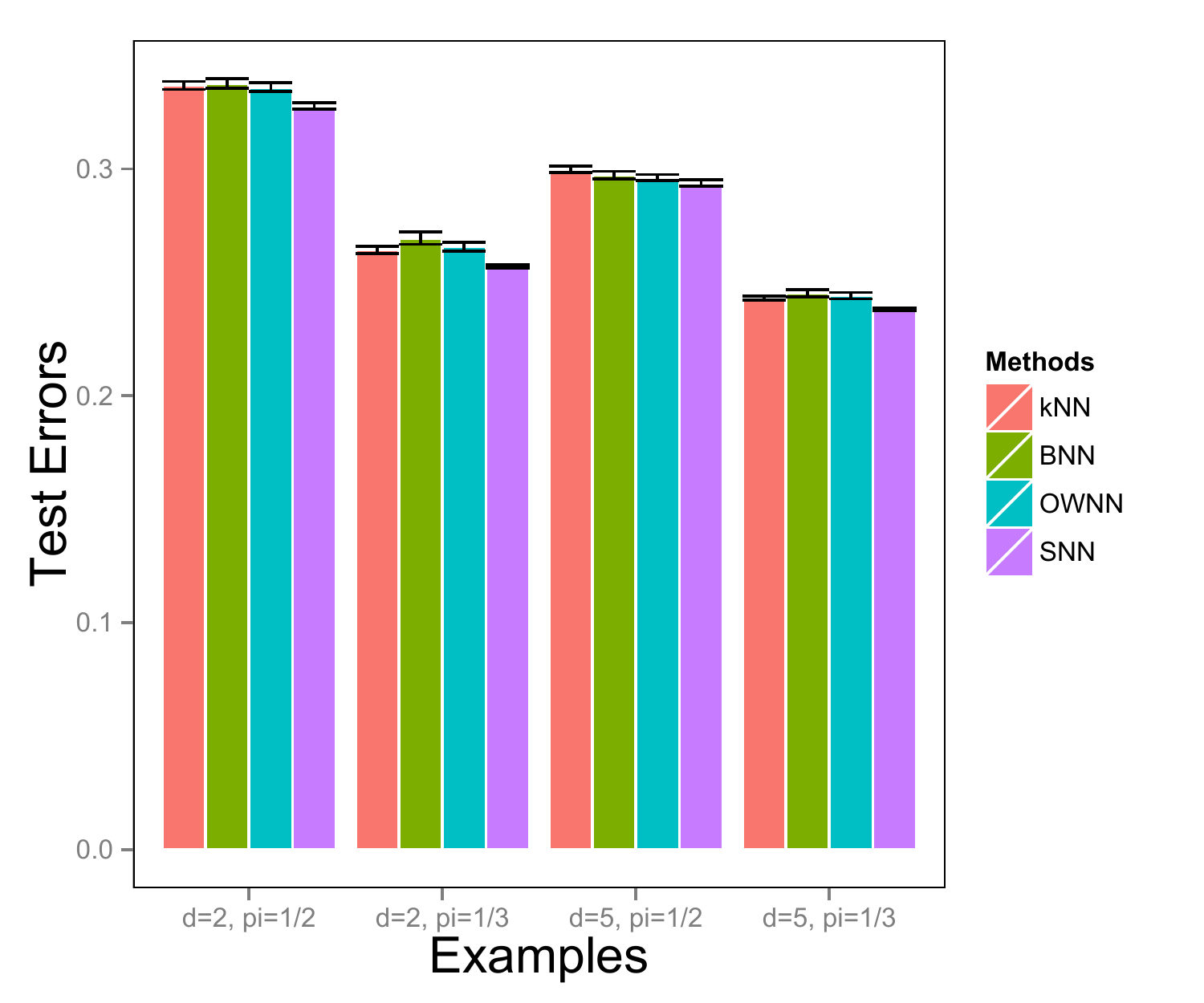}
\includegraphics[scale=0.5]{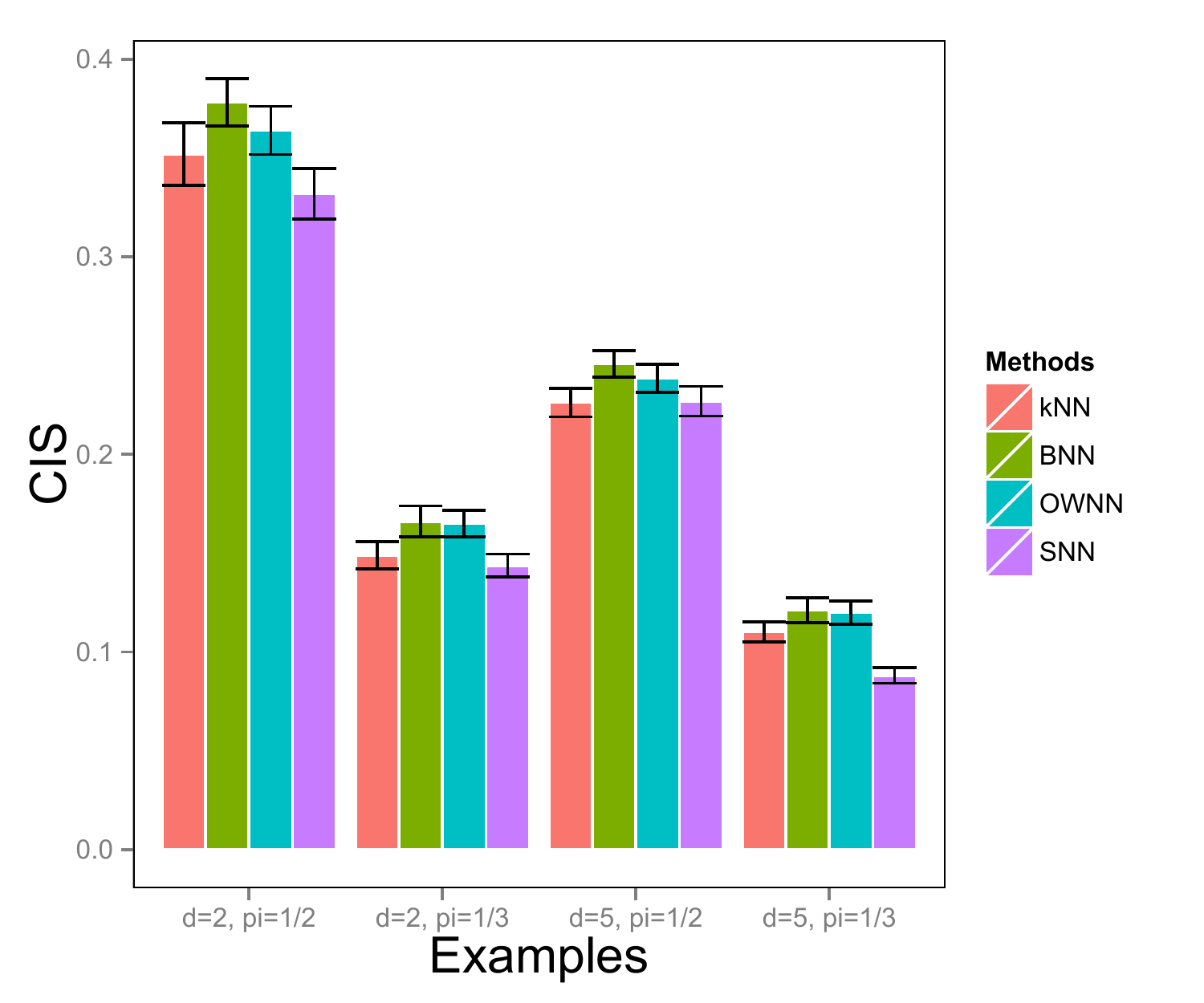}
\caption{\label{figure_sim3} \footnotesize Average test errors and CIS's (with standard error bar marked) of the $k$NN, BNN, OWNN, and SNN methods in Simulation 3. The ticks on the $x$-axis indicate the dimensions and prior class probability $\pi$ for different settings. Within each setting, the four methods are horizontally lined up (from the left are $k$NN, BNN, OWNN, and SNN).}
\end{center}
\end{figure}

Figures \ref{figure_sim2} and \ref{figure_sim3} summarize the results for Simulations 2 and 3. Similarly, in general, the difference in CIS is much obvious than the difference in the error. The SNN procedure obtains the minimal CIS in all 8 cases. Interestingly, the improvements are significant in all the four cases when $\pi_1 = 1/3$. Moreover, among 3 out of the 8 cases, our SNN achieves the smallest test errors and the improvements are significant. Even in cases where the error is not the smallest, the accuracy of SNN is close to the best classifier.

\subsection{Real Examples}
\label{real}
\begin{figure}[b!]
\begin{center}
\includegraphics[scale=0.5]{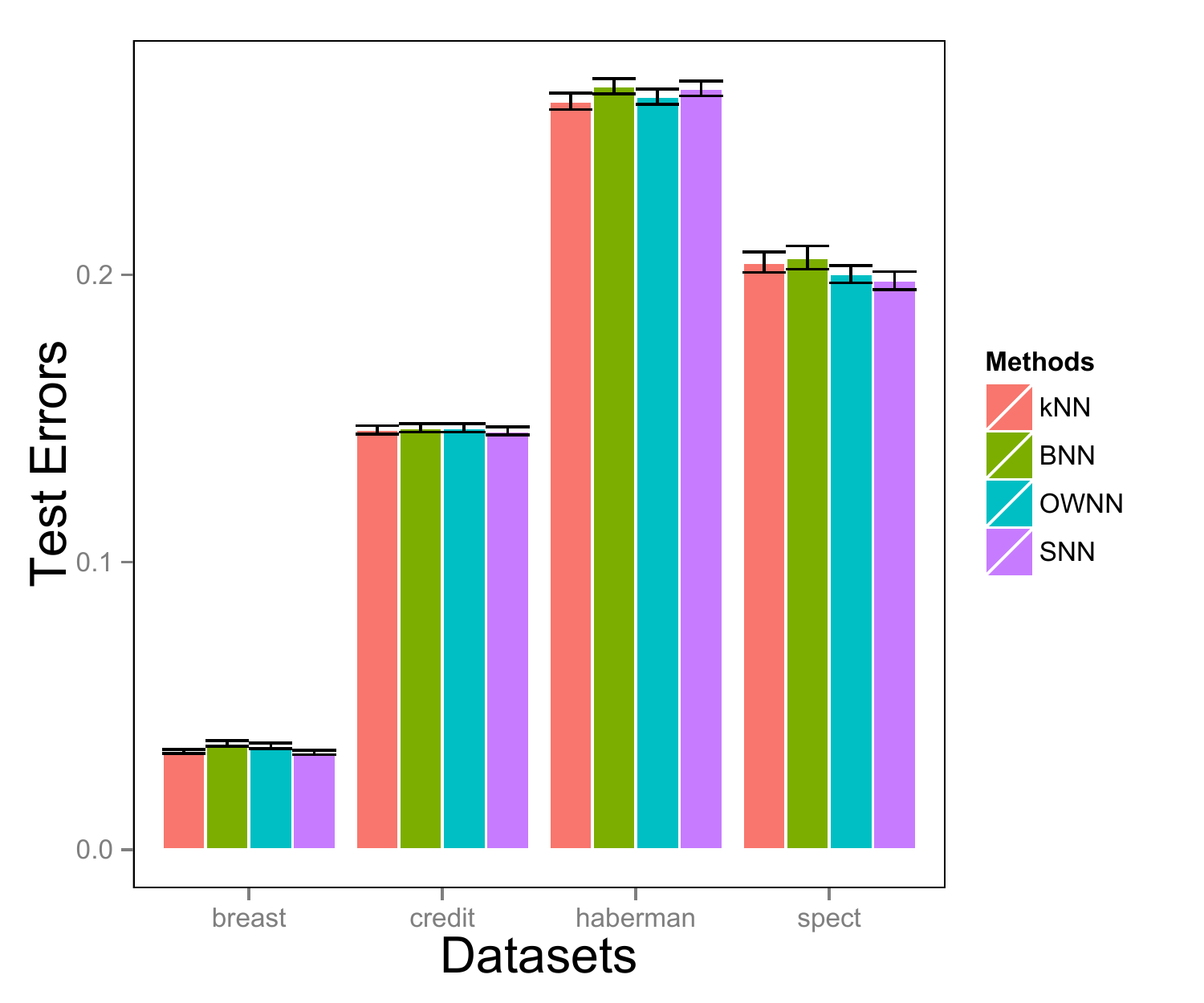}
\includegraphics[scale=0.5]{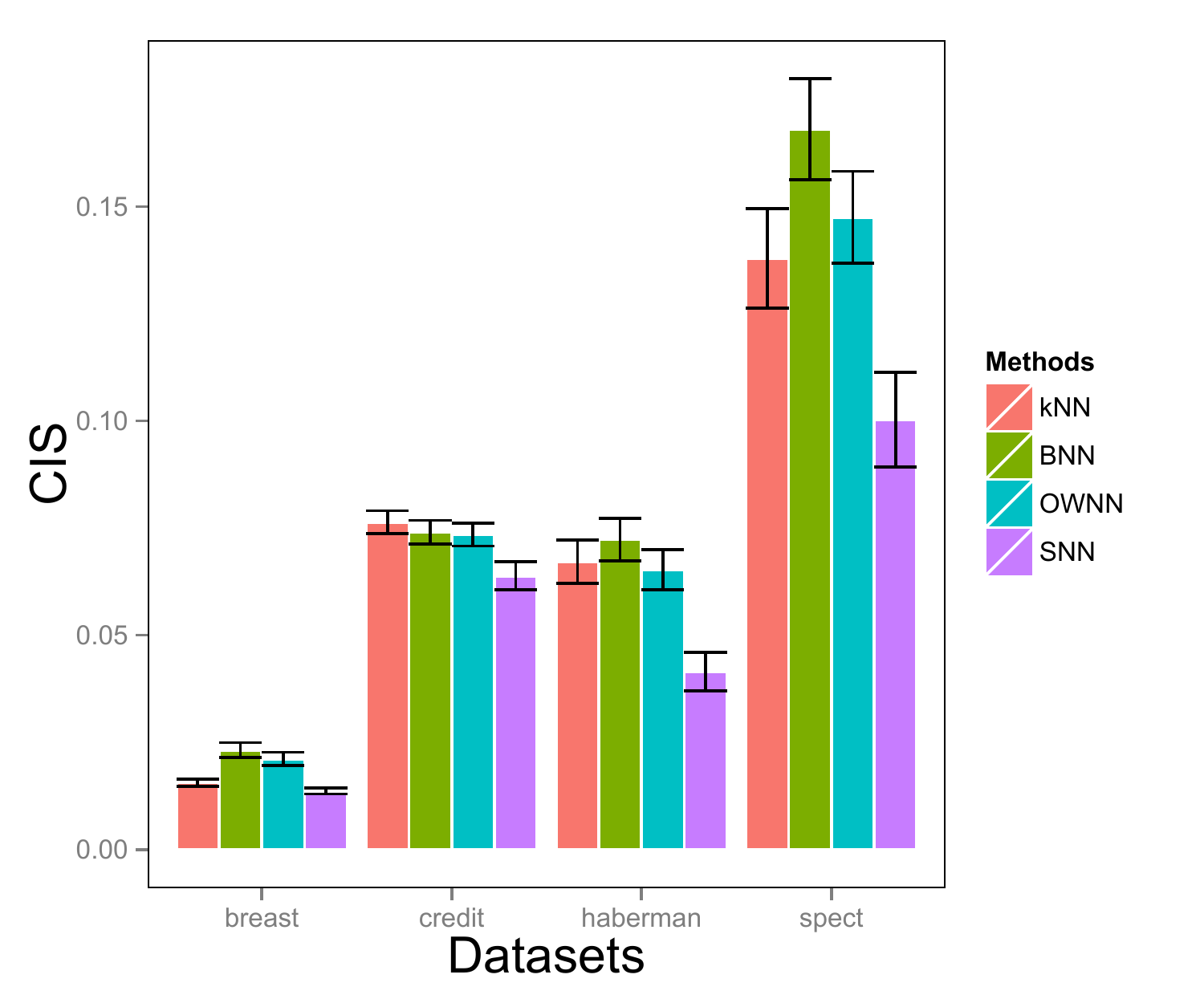}
\caption{\label{figure_real} \footnotesize Average test errors and CIS's (with standard error bar marked) of the $k$NN, BNN, OWNN and SNN methods for four data examples. The ticks on the $x$-axis indicate the names of the examples. Within each example, the four methods are horizontally lined up (from the left are $k$NN, BNN, OWNN, and SNN).}
\end{center}
\end{figure}

We extend the comparison to four real data sets publicly available in the UCI Machine Learning Repository \citep{BL13}. The first data set is the breast cancer data set ($breast$) collected by \citet{WM90}. There are 683 samples and 10 experimental measurement variables. The binary class label indicates whether the sample is benign or malignant. These 683 samples arrived periodically. In total, there are 8 groups of samples which reflect the chronological order of the data. A good classification procedure is expected to produce a stable classifier across these groups of samples. The second data set is the credit approval data set ($credit$). It consists of $690$ credit card applications and each application has $14$ attributes reflecting the user information. The binary class label refers to whether the application is positive or negative. The third data set is the haberman's survival data set ($haberman$) which contains $306$ cases from study conducted on the survival of patients who had undergone surgery for breast cancer. It has three attributes, age, patient's year of operation, and number of positive axillary nodes detected. The response variable indicates the survival status: either the patient survived 5 years or longer or the patient died within 5 years. The last data set is the SPECT heart data set ($spect$) which describes the diagnosing of cardiac Single Proton Emission Computed Tomography (SPECT) images. Each of the $267$ image sets (patients) had $22$ binary feature patterns and was classified into two classes: normal and abnormal.

We randomly split each data set into training and test sets with the equal size. The same tuning procedure as in the simulation is applied here. We compute the test error and the estimated CIS on the test set. The procedure is repeated 100 times and the average error and  CIS are reported in Figure \ref{figure_real}.

Similar to the simulation results, the SNN procedure obtains the minimal CIS in all four real data sets and the improvements in CIS are significant. The errors of OWNN and SNN have no significant difference, although OWNN is theoretically better in accuracy. These real experiments further illustrate that, with almost the same classification accuracy, our SNN procedure can achieve a significant improvement in the stability, which promotes the reproducibility.

\section{Conclusion}
\label{discussion}
Stability is an important and desirable property of a statistical procedure. It provides a foundation for the reproducibility, and reflects the credibility of those who use the procedure. To our best knowledge, our work is the first to propose a measure to quantify classification instability. The proposed SNN classification procedure enjoys increased classification stability with comparable classification accuracy to OWNN.

For classification problems, the classification accuracy is a primary concern, while stability is secondary. In many real cases, however, different classifiers may enjoy a comparable classification accuracy, and a classifier with a better stability stands out. The observation that our method can improve stability while maintaining the similar accuracy suggests that there may exist much more room for improving stability than for improving accuracy. This may be explained by the faster convergence rate of the regret than that of the CIS (Theorem \ref{thm:upperCISsnn}).

In theory, our SNN is shown to achieve the minimax optimal convergence rate in regret and a sharp convergence rate in CIS. Extensive experiments illustrate that SNN attains a significant improvement of stability over existing nearest neighbor classifiers, and sometimes even improves the accuracy. We implement the algorithm in a publicly available R package \texttt{snn}.

Our proposed SNN method is motivated by an asymptotic expansion of the CIS. Such a nice property may not exist for other more general classification methods. Hence, it is unclear how the stabilization idea can be carried over to other classifiers in a similar manner. That being said, the CIS measure can be used as a criterion for tuning parameter selection. There exists work in the literature which uses variable selection stability to select tuning parameter \citep{SWF13}. Classification stability and variable selection stability complement each other and can provide a comprehensive description of the reliability of a statistical procedure.  

For simplicity, we focus on the binary classification in this article. The generalization of the SNN classifier to multi-category classification problems \citep{LLW04, LS06, LY11} is an interesting topic to pursue in the future. Moreover, stability for the high-dimensional, low-sample size data is another important topic. Furthermore, in analyzing a big data set, a popular scheme is divide-and-conquer. It is an interesting research question on how to divide the data and choose the parameter wisely to ensure the optimal stability of a combined classifier.

\section*{Appendices}
\setcounter{subsection}{0}
\renewcommand{\thesubsection}{A.\Roman{subsection}}
\setcounter{equation}{0}
\renewcommand{\theequation}{A.\arabic{equation}}
\setcounter{lemma}{0}
\renewcommand{\thelemma}{A.\arabic{lemma}}

\subsection{Assumptions (A1) - (A4)}
\label{sec:assumptions}

For a smooth function $g$, we denote $\dot{g}(x)$ as its gradient vector at $x$. We assume the following conditions through all the article.

(A1) The set ${\cal R}\subset \mathbb R^d$ is a compact $d$-dimensional manifold with boundary $\partial{\cal R}$.

(A2) The set ${\cal S}=\{x\in {\cal R}: \eta(x)=1/2\}$ is nonempty. There exists an open subset $U_0$ of ${\mathbb R}^d$ which contains ${\cal S}$ such that: (i) $\eta$ is continuous on $U\backslash U_0$ with $U$ an open set containing ${\cal R}$; (ii) the restriction of the conditional distributions of $X$, $P_1$ and $P_2$, to $U_0$ are absolutely continuous with respect to Lebesgue measure, with twice continuously differentiable Randon-Nikodym derivatives $f_1$ and $f_2$.

(A3) There exists $\rho >0$ such that $\int_{{\mathbb R}^d}\|x\|^{\rho} d\bar{P}(x) < \infty$. Moreover, for sufficiently small $\delta>0$, $\inf_{x\in {\cal R}}\bar{P}(B_{\delta}(x))/(a_d\delta^d) \ge C_3 >0$, where $a_d=\pi^{d/2}/\Gamma(1+d/2)$, $\Gamma(\cdot)$ is gamma function, and $C_3$ is a constant independent of $\delta$.

(A4) For all $x\in {\cal S}$, we have $\dot{\eta}(x)\ne 0$, and for all $x\in {\cal S}\cap \partial{\cal R}$, we have $\dot{\partial \eta}(x)\ne 0$, where $\partial \eta$ is the restriction of $\eta$ to $\partial {\cal R}$. \hfill $\blacksquare$

\begin{remark}
Assumptions (A1)--(A4) have also been employed to show the asymptotic expansion of the regret of the $k$NN classifier \citep{HPS08}. The condition $\dot{\eta}(x)\ne 0$ in (A4) is equivalent to the margin condition with $\alpha=1$; see (2.1) in \citet{S12}. Furthermore, these assumptions ensure that $\bar{f}(x_0)$ and $\dot{\eta}(x_0)$ are bounded away from zero and infinity on ${\cal S}$.
\end{remark}

\subsection{Definitions of $a(x)$, $B_1$, $B_2$, and $W_{n,\beta}$}\label{sec:defwnb}

For a smooth function $g$: $\mathbb{R}^d\rightarrow \mathbb{R}$, let $g_j(x)$ its $j$th partial derivative at $x$, $\ddot{g}(x)$ the Hessian matrix at $x$, and $g_{jk}(x)$ the $(j,k)$th element of $\ddot{g}(x)$. Let $c_{j,d}=\int_{v:\|v\|\le 1} v_j^2 dv$. Define
$$
a(x)=\sum_{j=1}^d  \frac{c_{j,d}\{\eta_{j}(x)\bar{f}_j(x) + 1/2 \eta_{jj}(x)\bar{f}(x)\}}{a_d^{1+2/d} \bar{f}(x)^{1+2/d}}.
$$

We further define two distribution-related constants
\begin{eqnarray*}
B_1 = \int_{\cal S} \frac{\bar{f}(x)}{4\|\dot{\eta}(x)\|} d \textrm{Vol}^{d-1}(x),\quad B_2 = \int_{\cal S} \frac{\bar{f}(x)}{\|\dot{\eta}(x)\|} a(x)^2 d \textrm{Vol}^{d-1}(x),
\end{eqnarray*}
where $\textrm{Vol}^{d-1}$ is the natural $(d-1)$-dimensional volume measure that ${\cal S}$ inherits with ${\cal S}$ defined in Appendix \ref{sec:assumptions}. Based on Assumptions (A1)-(A4) in Appendix \ref{sec:assumptions}, $B_1$ and $B_2$ are finite with $B_1>0$ and $B_2\ge 0$, where $B_2=0$ only when $a(x)$ equals zero on ${\cal S}$.

In addition, for $\beta>0$, we denote $W_{n,\beta}$ as the set of $\bw_n$ satisfying (w.1)--(w.5).

(w.1) $\sum_{i=1}^n w_{ni}^2 \le n^{-\beta}$,

(w.2) $n^{-4/d}(\sum_{i=1}^n\alpha_iw_{ni})^2\le n^{-\beta}$, where $\alpha_i=i^{1+\frac{2}{d}}-(i-1)^{1+\frac{2}{d}}$,

(w.3) $n^{2/d}\sum_{i=k_2+1}^n w_{ni}/\sum_{i=1}^n \alpha_iw_{ni}\le 1/\log n$ with $k_2=\lfloor n^{1-\beta} \rfloor$,

(w.4) $\sum_{i=k_2+1}^n w_{ni}^2/\sum_{i=1}^n w_{ni}^2 \le 1/\log n$,

(w.5) $\sum_{i=1}^n w_{ni}^3/(\sum_{i=1}^n w_{ni}^2)^{3/2} \le 1/\log n$.

For the $k$NN classifier with $w_{ni}=k^{-1}\ind{1\le i\le k}$, (w.1)--(w.5) reduce to $\max(n^{\beta}, (\log n)^2) \le k \le \min(n^{(1-\beta d /4)}, n^{1-\beta})$. See \citet{S12} for a detailed discussion of $W_{n,\beta}$.   \hfill $\blacksquare$

\subsection{Proof of Theorem \ref{thm:CIS}}
\label{sec:proofcis}
Note that $\textrm{CIS}(\textrm{WNN})= {\mathbb P}_{{\cal D}_1,{\cal D}_2,X}\Big(\widehat{\phi}_{n1}^{\bw_n}(X) \ne \widehat{\phi}_{n2}^{\bw_n}(X)\Big)$ can be expressed in the following way.
\begin{eqnarray*}
&&\textrm{CIS}(\textrm{WNN}) \\
&=& {\mathbb E}_{X} \Big[ {\mathbb P}_{{\cal D}_1,{\cal D}_2}\Big(\widehat{\phi}_{{\cal D}_1}^{\bw_n}(X) \ne \widehat{\phi}_{{\cal D}_2}^{\bw_n}(X)\Big|X\Big) \Big]\\
&=& {\mathbb E}_{X} \Big[ {\mathbb P}_{{\cal D}_1,{\cal D}_2}\Big(\widehat{\phi}_{{\cal D}_1}^{\bw_n}(X)=1, \widehat{\phi}_{{\cal D}_2}^{\bw_n}(X) =2 \Big|X \Big)\Big] + {\mathbb E}_{X} \Big[ {\mathbb P}_{{\cal D}_1,{\cal D}_2}\Big(\widehat{\phi}_{{\cal D}_1}^{\bw_n}(X)=2, \widehat{\phi}_{{\cal D}_2}^{\bw_n}(X) =1 \Big|X\Big)\Big] \\
&=& {\mathbb E}_{X} \Big[ 2 {\mathbb P}_{{\cal D}_1}\Big(\widehat{\phi}_{{\cal D}_1}^{\bw_n}(X)=1|X\Big) \Big(1- {\mathbb P}_{{\cal D}_1}\Big(\widehat{\phi}_{{\cal D}_1}^{\bw_n}(X)=1|X\Big)\Big) \Big],
\end{eqnarray*}
where the last equality is valid because ${\cal D}_1$ and ${\cal D}_2$ are i.i.d. samples. Without loss of generality, we consider a generic sample ${\cal D}=\{(X_i,Y_i),i=1,\ldots,n\}$. Given $X=x$, we define $(X_{(i)},Y_{(i)})$ such that $\|X_{(1)}-x\|\le \|X_{(2)}-x\|\le \ldots \le \|X_{(n)}-x\|$ with $\|\cdot\|$ the Euclidean norm. Denote the estimated regression function $S_n(x) = \sum_{i=1}^n w_{ni}\ind{Y_{(i)}=1}$. We have
\begin{align*}
{\mathbb E}_{X} \Big[ {\mathbb P}\Big(\widehat{\phi}_{{\cal D}}^{\bw_n}(X)=1|X\Big)\Big] &= \int_{{\cal R}} {\mathbb P}\Big(S_n(x) \ge 1/2 \Big) d \bar{P}(x),\\
{\mathbb E}_{X} \Big[{\mathbb P}^2\Big(\widehat{\phi}_{{\cal D}}^{\bw_n}(X)=1|X\Big)\Big] &= \int_{{\cal R}} {\mathbb P}^2\Big(S_n(x) \ge 1/2 \Big) d \bar{P}(x),
\end{align*}
where $\bar{P}(x)$ is the marginal distribution of $X$. For the sake of simplicity, $\mathbb P$ denotes the probability with respect to $\cal D$. Hence, CIS satisfies
\begin{align*}
\textrm{CIS}(\textrm{WNN})/2 &= \int_{{\cal R}} {\mathbb P}(S_n(x) \ge 1/2 ) \Big(1- {\mathbb P}(S_n(x) \ge 1/2)\Big)d \bar{P}(x)\\
&=  \int_{{\cal R}} \left\{ {\mathbb P}(S_n(x) < 1/2 ) - \ind{\eta(x)<1/2}\right\} d\bar{P}(x) \nonumber\\
&~~~ -  \int_{{\cal R}} \left\{  {\mathbb P}^2(S_n(x) < 1/2) - \ind{\eta(x)<1/2} \right\}d \bar{P}(x)
\end{align*}

Denote the boundary ${\cal S}=\{x\in {\cal R}: \eta(x)=1/2\}$. For $\epsilon>0$, let ${\cal S}^{\epsilon\epsilon} = \{x\in {\mathbb R}^d: \eta(x)=1/2 ~\textrm{and}~ \textrm{dist}(x, {\cal S})<\epsilon \}$, where $\textrm{dist}(x, {\cal S})=\inf_{x_0\in {\cal S}} \|x-x_0\|$. We will focus on the set
$$
{\cal S}^{\epsilon} = \left\{x_0 + t\frac{\dot{\eta}(x_0)}{\|\dot{\eta}(x_0)\|}: x_0 \in {\cal S}^{\epsilon\epsilon}, |t| < \epsilon \right\}.
$$

Let $\mu_n(x)={\mathbb E}\{S_n(x)\}$, $\sigma_n^2(x)=\textrm{Var}\{S_n(x)\}$, and $\epsilon_n=n^{-\beta d/4}$. Denote $s_n^2=\sum_{i=1}^n w_{ni}^2$ and $t_n=n^{-2/d}\sum_{i=1}^n \alpha_i w_{ni}$. \citet{S12} showed that, uniformly for $\bw_n\in W_{n,\beta}$,
\begin{eqnarray}
\sup_{x\in {\cal S}^{\epsilon_n}} |\mu_n(x) - \eta(x) - a(x)t_n| &=& o(t_n),\label{step1:tn}\\
\sup_{x\in {\cal S}^{\epsilon_n}} \left|\sigma_n^2(x)-\frac{1}{4}s_n^2\right| &=& o(s_n^2). \label{step1:sn}
\end{eqnarray}

We organize our proof in three steps. In Step 1, we focus on analyzing on the set ${\cal R} \cap {\cal S}^{\epsilon_n} $; in Step 2, we focus on the complement set ${\cal R} \backslash {\cal S}^{\epsilon_n} $; Step 3 combines the results and applies a normal approximation to yield the final conclusion.

{\it Step 1}: For $x_0\in {\cal S}$ and $t\in {\mathbb R}$, denote $x_0^t=x_0+t \dot{\eta}(x_0)/\|\dot{\eta}(x_0)\|$. Denote $\bar{f}=\pi_1 f_1 + (1-\pi_1) f_2$ as the Radon-Nikodym derivative with respect to Lebesgue measure of the restriction of $\bar{P}$ to ${\cal S}^{\epsilon_n}$ for large $n$. We need to show that, uniformly for $\bw_n\in W_{n,\beta}$,
\begin{eqnarray}
&&\int_{{\cal R}\cap{{\cal S}^{\epsilon_n}}} \left\{ {\mathbb P}(S_n(x) < 1/2) - \ind{\eta(x)<1/2} \right\} d \bar{P}(x) = \nonumber\\
&&~~~\int_{{\cal S}} \int_{-\epsilon_n}^{\epsilon_n} {\bar f}(x_0^t) \left\{{\mathbb P}\Big(S_n(x_0^t) < 1/2\Big) - \ind{t<0}\right\} dt d\textrm{Vol}^{d-1}(x_0)\{1+o(1)\};\label{tube}\\
&&\int_{{\cal R}\cap{{\cal S}^{\epsilon_n}}} \left\{ {\mathbb P}^2(S_n(x) < 1/2) - \ind{\eta(x)<1/2} \right\} d \bar{P}(x) = \nonumber\\
&&~~~\int_{{\cal S}} \int_{-\epsilon_n}^{\epsilon_n} {\bar f}(x_0^t) \left\{{\mathbb P}^2\Big(S_n(x_0^t) < 1/2\Big) - \ind{t<0}\right\} dt d\textrm{Vol}^{d-1}(x_0)\{1+o(1)\}.\label{tube2}
\end{eqnarray}

According to \citet{S12}, for large $n$, we define the map $\phi(x_0,t\frac{\dot{\eta}(x_0)}{\|\dot{\eta}(x_0)\|})=x_0^t$, and note that
$$
\det \dot{\phi}\Big(x_0,t\frac{\dot{\eta}(x_0)}{\|\dot{\eta}(x_0)\|}\Big)dt d\textrm{Vol}^{d-1}(x_0) = \{1+o(1)\}dt d\textrm{Vol}^{d-1}(x_0),
$$
uniformly in $(x_0, t \dot{\eta}(x_0)/\|\dot{\eta}(x_0)\|)$ for $x_0\in {\cal S}$ and $|t|<\epsilon_n$, where $\det$ is the determinant. Then the theory of integration on manifolds \citep{G04} implies that, uniformly for $\bw_n\in W_{n,\beta}$,
\begin{eqnarray}
&&\int_{{\cal S}^{\epsilon_n}} \left\{ {\mathbb P}(S_n(x) < 1/2) - \ind{\eta(x)<1/2} \right\} d \bar{P}(x) = \nonumber\\
&&\int_{{\cal S}^{\epsilon_n\epsilon_n}} \int_{-\epsilon_n}^{\epsilon_n} {\bar f}(x_0^t) \left\{{\mathbb P}\Big(S_n(x_0^t) < 1/2\Big) - \ind{t<0}\right\} dt d\textrm{Vol}^{d-1}(x_0)\{1+o(1)\}.\nonumber
\end{eqnarray}
Furthermore, we can replace ${\cal S}^{\epsilon_n}$ with ${\cal R}\cap{{\cal S}^{\epsilon_n}}$ since ${\cal S}^{\epsilon_n}\backslash {\cal R}\subseteq \{x\in \mathbb{R}^d: \textrm{dist}(x,\partial {\cal S})<\epsilon_n\}$ and the latter has volume $O(\epsilon_n^2)$ by Weyl's tube formula \citep{G04}. Similarly, we can safely replace ${\cal S}^{\epsilon_n\epsilon_n}$ with ${\cal S}$. Therefore, $(\ref{tube})$ holds. Similar arguments imply $(\ref{tube2})$.

{\it Step 2}: Bound the contribution to CIS from ${\cal R}\backslash {\cal S}^{\epsilon_n}$. We show that, for all $M>0$,
\begin{eqnarray}
\sup_{\bw_n\in W_{n,\beta}} \int_{{\cal R}\backslash {\cal S}^{\epsilon_n}} \left\{ {\mathbb P}\Big(S_n(x) < 1/2\Big) - \ind{\eta(x)<1/2} \right\} d \bar{P}(x) &=& O(n^{-M}),\label{bound1}\\
\sup_{\bw_n\in W_{n,\beta}} \int_{{\cal R}\backslash {\cal S}^{\epsilon_n}} \left\{ {\mathbb P}^2\Big(S_n(x) < 1/2\Big) - \ind{\eta(x)<1/2} \right\} d \bar{P}(x) &=& O(n^{-M}).\label{bound2}
\end{eqnarray}
Here $(\ref{bound1})$ follows from the fact $|{\mathbb P}(S_n(x) < \frac{1}{2}) - \ind{\eta(x)<1/2}| = O(n^{-M})$ for all $M>0$, uniformly for $\bw_n\in W_{n,\beta}$ and $x\in {\cal R}\backslash {\cal S}^{\epsilon_n}$ \citep{S12}. Furthermore, $(\ref{bound2})$ holds since
$$
\Big|{\mathbb P}^2\Big(S_n(x) < 1/2\Big) - \ind{\eta(x)<1/2}\Big|\le 2\Big|{\mathbb P}\Big(S_n(x) < 1/2\Big) - \ind{\eta(x)<1/2}\Big|.
$$

{\it Step 3}: In the end, we will show
\begin{eqnarray}
&&\int_{{\cal S}} \int_{-\epsilon_n}^{\epsilon_n} {\bar f}(x_0^t) \left\{{\mathbb P}\Big(S_n(x_0^t) < 1/2\Big) - \ind{t<0}\right\} dt d\textrm{Vol}^{d-1}(x_0) \nonumber\\
&&~~~-\int_{{\cal S}} \int_{-\epsilon_n}^{\epsilon_n} {\bar f}(x_0^t) \left\{{\mathbb P}^2\Big(S_n(x_0^t) < 1/2\Big) - \ind{t<0}\right\} dt d\textrm{Vol}^{d-1}(x_0) \nonumber\\
&=& \frac{1}{2}B_3 s_n +  o(s_n+t_n).
\end{eqnarray}

We first apply the nonuniform version of Berry-Esseen Theorem to approximate ${\mathbb P}(S_n(x_0^t) < 1/2)$. Let $Z_i= (w_{ni}\ind{Y_{(i)}=1} - w_{ni} \mathbb{E} [\ind{Y_{(i)}=1}])/\sigma_n(x)$ and $W=\sum_{i=1}^n Z_i$. Note that $\mathbb{E}(Z_i)=0$, $\textrm{Var}(Z_i)<\infty$, and $\textrm{Var}(W)=1$. Then the nonuniform Berry-Esseen Theorem \citep{B77} implies that
$$
\Big|\mathbb{P}(W\le y) - \Phi(y)\Big| \le \frac{M_1}{n^{1/2}( 1 + |y|^3)},
$$
where $\Phi$ is the standard normal distribution function and $M_1$ is a constant. Therefore,
\begin{equation}
\sup_{x_0\in {\cal S}}\sup_{t\in [-\epsilon_n,\epsilon_n]} \left|\mathbb{P}\Big(\frac{S_n(x_0^t)-\mu_n(x_0^t)}{\sigma_n(x_0^t)}\le y \Big) - \Phi(y) \right| \le \frac{M_1}{n^{1/2} (1 + |y|^3)}. \label{BE_thm}
\end{equation}
Thus, we have
\begin{eqnarray}
&&\int_{{\cal S}} \int_{-\epsilon_n}^{\epsilon_n} {\bar f}(x_0^t) \left\{{\mathbb P}\Big(S_n(x_0^t) < 1/2\Big) - \ind{t<0}\right\} dt d\textrm{Vol}^{d-1}(x_0) \nonumber\\
&&~~~~~~= \int_{{\cal S}} \int_{-\epsilon_n}^{\epsilon_n} \bar{f}(x_0^t) \left\{\Phi\Big(\frac{1/2 - \mu_n(x_0^t)}{\sigma_n(x_0^t)} \Big) - \ind{t<0}\right\} dt d\textrm{Vol}^{d-1}(x_0) + o(s_n^2+t_n^2), \nonumber
\end{eqnarray}
where the remainder term $o(s_n^2+t_n^2)$ is due to $(\ref{BE_thm})$ by slightly modifying the proof of A.21 in \citet{S12}.

Furthermore, Taylor expansion leads to
$$
\bar{f}(x_0^t) = \bar{f}(x_0) + (\dot{\bar{f}}(x_0))^T\frac{\dot{\eta}(x_0)}{\|\dot{\eta}(x_0)\|} t + o(t).
$$
Therefore,
\begin{eqnarray}
&&\int_{{\cal S}} \int_{-\epsilon_n}^{\epsilon_n} {\bar f}(x_0^t) \left\{{\mathbb P}\Big(S_n(x_0^t) < 1/2\Big) - \ind{t<0}\right\} dt d\textrm{Vol}^{d-1}(x_0) \label{exp1}\\
&=& \int_{{\cal S}} \int_{-\epsilon_n}^{\epsilon_n} \bar{f}(x_0) \left\{\Phi\Big(\frac{-2t\|\dot{\eta}(x_0)\|- 2a(x_0)t_n}{s_n} \Big) - \ind{t<0}\right\} dt d\textrm{Vol}^{d-1}(x_0)\nonumber\\
&&+\int_{{\cal S}} \int_{-\epsilon_n}^{\epsilon_n} \frac{\dot{\bar{f}}(x_0)^T\dot{\eta}(x_0)t}{\|\dot{\eta}(x_0)\|} \left\{\Phi\Big(\frac{-2t\|\dot{\eta}(x_0)\|- 2a(x_0)t_n}{s_n} \Big) - \ind{t<0}\right\} dt d\textrm{Vol}^{d-1}(x_0) + R_1, \nonumber
\end{eqnarray}
where
\begin{align*}
R_1 &= \int_{{\cal S}} \int_{-\epsilon_n}^{\epsilon_n} \bar{f}(x_0) \left\{ \Phi\Big(\frac{1/2 - \mu_n(x_0^t)}{\sigma_n(x_0^t)}\Big)-  \Phi\Big(\frac{-2t\|\dot{\eta}(x_0)\|- 2a(x_0)t_n}{s_n} \Big)\right\} dt d\textrm{Vol}^{d-1}(x_0)\\
&~~~~ + \int_{{\cal S}} \int_{-\epsilon_n}^{\epsilon_n} \frac{\dot{\bar{f}}(x_0)^T\dot{\eta}(x_0)t}{\|\dot{\eta}(x_0)\|} \left\{ \Phi\Big(\frac{1/2 - \mu_n(x_0^t)}{\sigma_n(x_0^t)}\Big)-  \Phi\Big(\frac{-2t\|\dot{\eta}(x_0)\|- 2a(x_0)t_n}{s_n} \Big)\right\} dt d\textrm{Vol}^{d-1}(x_0)\\
&~~~~ + o(s_n^2+t_n^2)\\
&\buildrel \Delta \over = R_{11}+R_{12}+o(s_n^2+t_n^2).
\end{align*}

Next we show $R_1=o(s_n+t_n)$. Denote $r_{x_0}=\frac{-a(x_0)t_n}{\|\dot{\eta}(x_0)s_n\|}$. According to $(\ref{step1:tn})$ and $(\ref{step1:sn})$, for a sufficiently small $\epsilon\in (0,\inf_{x_0\in {\cal S}} \|\dot{\eta}(x_0)\|)$ and large $n$, for all $\bw_n\in W_{n,\beta}$, $x_0\in {\cal S}$ and $r\in[-\epsilon_n/s_n,\epsilon_n/s_n]$, \citet{S12} showed that
$$
\Big| \frac{1/2-\mu_n(x_0^{rs_n})}{\sigma_n(x_0^{rs_n})} -[-2\|\dot{\eta}(x_0)\|(r-r_{x_0})] \Big| \le \epsilon^2(|r|+t_n/s_n).
$$
In addition, when $|r-r_{x_0}|\le \epsilon t_n/s_n$,
\begin{eqnarray*}
\Big| \Phi\Big(\frac{1/2-\mu_n(x_0^{rs_n})}{\sigma_n(x_0^{rs_n})}\Big) -\Phi\Big(-2\|\dot{\eta}(x_0)\|(r-r_{x_0})\Big) \Big|\le 1
\end{eqnarray*}
and when $\epsilon t_n/s_n < |r| < t_n/s_n$,
\begin{eqnarray*}
\Big| \Phi\Big(\frac{1/2-\mu_n(x_0^{rs_n})}{\sigma_n(x_0^{rs_n})}\Big) -\Phi\Big(-2\|\dot{\eta}(x_0)\|(r-r_{x_0})\Big) \Big|\le \epsilon^2(|r|+t_n/s_n)\phi(\|\dot{\eta}(x_0)\||r-r_{x_0}|),
\end{eqnarray*}
where $\phi$ is the density function of standard normal distribution.

Therefore, we have
\begin{eqnarray}
|R_{11}| &\le & \int_{{\cal S}} \int_{-\epsilon_n}^{\epsilon_n} \bar{f}(x_0) \left| \Phi\Big(\frac{1/2 - \mu_n(x_0^t)}{\sigma_n(x_0^t)}\Big)-  \Phi\Big(\frac{-2t\|\dot{\eta}(x_0)\|- 2a(x_0)t_n}{s_n} \Big)\right| dt d\textrm{Vol}^{d-1}(x_0)\nonumber\\
&\le & \bar{f}(x_0) s_n \int_{|r-r_{x_0}|\le \epsilon t_n/s_n} dr + \bar{f}(x_0) s_n \epsilon^2 \int_{-\infty}^{\infty} (|r|+t_n/s_n)\phi(\|\dot{\eta}(x_0)\||r-r_{x_0}|) dr\nonumber\\
&\le& \epsilon (t_n + s_n) \label{R11}.
\end{eqnarray}

Similarly,
\begin{eqnarray*}
&&|R_{12}| \nonumber\\
&\le & \int_{{\cal S}} \int_{-\epsilon_n}^{\epsilon_n} \frac{\dot{\bar{f}}(x_0)^T\dot{\eta}(x_0)t}{\|\dot{\eta}(x_0)\|} \left| \Phi\Big(\frac{1/2 - \mu_n(x_0^t)}{\sigma_n(x_0^t)}\Big)-  \Phi\Big(\frac{-2t\|\dot{\eta}(x_0)\|- 2a(x_0)t_n}{s_n} \Big)\right| dt d\textrm{Vol}^{d-1}(x_0) \nonumber\\
&\le & \bar{f}(x_0) \epsilon s_n^2 \int_{|r-r_{x_0}|\le \epsilon t_n/s_n} |r| dr + \bar{f}(x_0) s_n^2 \epsilon^2 \int_{-\infty}^{\infty} (|r|+t_n/s_n)\phi(\|\dot{\eta}(x_0)\||r-r_{x_0}|) dr \nonumber\\
&\le& \epsilon (t_n^2 + s_n^2).
\end{eqnarray*}
The inequality above, along with with $(\ref{R11})$, leads to $R_1=o(s_n+t_n)$.

By similar arguments, we have
\begin{align}
&\int_{{\cal S}} \int_{-\epsilon_n}^{\epsilon_n} {\bar f}(x_0^t) \left\{{\mathbb P}^2\Big(S_n(x_0^t) < 1/2\Big) - \ind{t<0}\right\} dt d\textrm{Vol}^{d-1}(x_0) \label{exp2}\\
=& \int_{{\cal S}} \int_{-\epsilon_n}^{\epsilon_n} \bar{f}(x_0) \left\{\Phi^2\Big(\frac{-2t\|\dot{\eta}(x_0)\|- 2a(x_0)t_n}{s_n} \Big) - \ind{t<0}\right\} dt d\textrm{Vol}^{d-1}(x_0)\nonumber\\
&+\int_{{\cal S}} \int_{-\epsilon_n}^{\epsilon_n} \frac{\dot{\bar{f}}(x_0)^T\dot{\eta}(x_0)t}{\|\dot{\eta}(x_0)\|} \left\{\Phi^2\Big(\frac{-2t\|\dot{\eta}(x_0)\|- 2a(x_0)t_n}{s_n} \Big) - \ind{t<0}\right\} dt d\textrm{Vol}^{d-1}(x_0)\nonumber\\
&+ o(s_n+t_n). \nonumber
\end{align}

Finally, after substituting $t=us_n/2$ in (\ref{exp1}) and  (\ref{exp2}), we have, up to $o(s_n+t_n)$ difference,
\begin{eqnarray*}
&&\textrm{CIS}(\textrm{WNN})/2 \nonumber\\
&=&\frac{s_n}{2}\int_{{\cal S}} \int_{-\infty}^{\infty} \bar{f}(x_0) \left\{\Phi\Big(-\|\dot{\eta}(x_0)\|u-\frac{2a(x_0)t_n}{s_n} \Big) - \ind{u<0} \right\} du d\textrm{Vol}^{d-1}(x_0) \nonumber\\
&&+ \frac{s_n^2}{4}\int_{{\cal S}} \int_{-\infty}^{\infty} \frac{(\dot{\bar{f}}(x_0))^T\dot{\eta}(x_0)}{\|\dot{\eta}(x_0)\|} u \left\{\Phi\Big(-\|\dot{\eta}(x_0)\|u-\frac{2a(x_0)t_n}{s_n} \Big) - \ind{u<0}\right\} dt d\textrm{Vol}^{d-1}(x_0) \nonumber\\
&&- \frac{s_n}{2}\int_{{\cal S}} \int_{-\infty}^{\infty} \bar{f}(x_0) \left\{\Phi^2\Big(-\|\dot{\eta}(x_0)\|u-\frac{2a(x_0)t_n}{s_n} \Big) - \ind{u<0}\right\} du d\textrm{Vol}^{d-1}(x_0) \nonumber\\
&& - \frac{s_n^2}{4}\int_{{\cal S}} \int_{-\infty}^{\infty} \frac{(\dot{\bar{f}}(x_0))^T\dot{\eta}(x_0)}{\|\dot{\eta}(x_0)\|} u \left\{\Phi^2\Big(-\|\dot{\eta}(x_0)\|u-\frac{2a(x_0)t_n}{s_n} \Big) - \ind{u<0}\right\} dt d\textrm{Vol}^{d-1}(x_0) \nonumber\\
&=& I + II - III - IV.
\end{eqnarray*}

According to Lemma \ref{lemma}, we have
\begin{eqnarray*}
I - III &=& \left [\int_{\cal S} \frac{{\bar{f}}(x_0)}{2\sqrt{\pi}\|\dot{\eta}(x_0)\|}  d\textrm{Vol}^{d-1}(x_0) \right] s_n = \frac{1}{2}B_3 s_n\\
II - IV &=& - \left [\int_{\cal S} \frac{(\dot{\bar{f}}(x_0))^T\dot{\eta}(x_0)a(x_0)}{2\sqrt{\pi}(\|\dot{\eta}(x_0)\|)^3}  d\textrm{Vol}^{d-1}(x_0)\right] s_nt_n = \frac{1}{2}B_4s_nt_n.
\end{eqnarray*}
Therefore, the desirable result is obtained by noting that $B_4 s_nt_n = o(s_n+t_n)$. This concludes the proof of Theorem \ref{thm:CIS}. \hfill $\blacksquare$

\subsection{Proof of Theorem \ref{thm:optimal}} For any weight $\bw_n$, the Lagrangian of $(\ref{formal})$ is
$$
L(\bw_n)=\Big(\sum_{i=1}^n \frac{\alpha_i w_{ni}}{n^{2/d}}\Big)^2 + \lambda \sum_{i=1}^n w_{ni}^2 + \nu(\sum_{i=1}^{n}w_{ni}-1).
$$
Considering the constraint of nonnegative weights, we denote $k^{*}=\max\{i:w_{ni}^{*}>0\}$. Setting derivative of $L(\bw_n)$ to be $0$, we have
\begin{equation}
\frac{\partial L(\bw_n)}{\partial w_{ni}} = 2 n^{-4/d}\alpha_i\sum_{i=1}^{k^*} \alpha_i w_{ni} + 2\lambda w_{ni} +  \nu =0. \label{derivative}
\end{equation}
Summing $(\ref{derivative})$ from 1 to $k^*$, and multiplying $(\ref{derivative})$ by $\alpha_i$ and then summing from 1 to $k^*$ yields
\begin{eqnarray*}
2 n^{-4/d}(k^*)^{1+2/d}\sum_{i=1}^{k^*}\alpha_i w_{ni}  + 2\lambda + \nu k^* &=& 0\\
2 n^{-4/d}\sum_{i=1}^{k^*}\alpha_i w_{ni} \sum_{i=1}^{k^*} \alpha_i^2 + 2\lambda \sum_{i=1}^{k^*}\alpha_i w_{ni} + \nu (k^*)^{1+2/d} &=& 0.
\end{eqnarray*}
Therefore, we have
\begin{equation}
w_{ni}^{*} = \frac{1}{k^*} + \frac{(k^*)^{4/d}-(k^*)^{2/d}\alpha_i}{\sum_{i=1}^{k^*} \alpha_i^2+\lambda n^{4/d} -(k^*)^{1+4/d}} \label{optweight0}
\end{equation}
Here $w_{ni}^{*}$ is decreasing in $i$ since $\alpha_i$ is increasing in $i$ and $\sum_{i=1}^{k^{*}}\alpha_i^2>(k^{*})^{1+4/d}$ from Lemma \ref{alpha}. Next we solve for $k^*$. According to the definition of $k^{*}$, we only need to find $k$ such that $w_{nk}^{*}=0$. Using the results from Lemma \ref{alpha}, solving this equation reduces to solving $k^*$ such that
$$
(1+\frac{2}{d})(k^*-1)^{2/d} \le \lambda n^{4/d}(k^*)^{-1-2/d} + \frac{(d+2)^2}{d(d+4)}(k^*)^{2/d}\{1+O(\frac{1}{k^*})\} \le (1+\frac{2}{d})(k^*)^{2/d}.
$$
Therefore, for large $n$, we have
\begin{equation}
k^*= \Big\lfloor \Big\{\frac{d(d+4)}{2(d+2)}\Big\}^{\frac{d}{d+4}}\lambda^{\frac{d}{d+4}}n^{\frac{4}{d+4}}\Big\rfloor.\nonumber
\end{equation}
Plugging $k^*$ and the result $(\ref{sum})$ in Supplementary into $(\ref{optweight0})$ yields the optimal weight. \hfill $\blacksquare$

\subsection{Proof of Theorem \ref{thm:upperCIS}}
\label{sec:proofuppercis}
Following the proofs of Lemma 3.1 in \citet{AT07}, we consider the sets $A_j \subset {\cal R}$
\begin{eqnarray*}
A_0 &=& \{x\in {\cal R}:  0<|\eta(x)-1/2|\le \delta\},\\
A_j &=& \{x\in {\cal R}:  2^{j-1}\delta < |\eta(x)-1/2|\le 2^j\delta\} \textrm{~for~} j\ge 1.
\end{eqnarray*}

For the classification procedure $\Psi(\cdot)$, we have
\begin{eqnarray*}
\textrm{CIS}(\Psi) = {\mathbb E}[\ind{\widehat{\phi}_{n1}(X) \ne \widehat{\phi}_{n2}(X)}],
\end{eqnarray*}
where $\widehat{\phi}_{n1}$ and $\widehat{\phi}_{n2}$ are classifiers obtained by applying $\Psi(\cdot)$ to two independently and identically distributed samples ${\cal D}_1$ and ${\cal D}_2$, respectively. Denote the Bayes classifier $\phi^{\textrm{Bayes}}$, we have
\begin{eqnarray*}
\textrm{CIS}(\Psi) &=& 2 {\mathbb E}[\ind{\widehat{\phi}_{n1}(X) = \phi^{\textrm{Bayes}}(X), \widehat{\phi}_{n2}(X)\ne \phi^{\textrm{Bayes}}(X)}]\\
&=& 2 {\mathbb E}[ \{1 - \ind{\widehat{\phi}_{n1}(X) \ne \phi^{\textrm{Bayes}}(X)}\}  \ind{\widehat{\phi}_{n2}(X)\ne \phi^{\textrm{Bayes}}(X)}] \\
&=& 2 {\mathbb E}_X[ {\mathbb P}_{{\cal D}_1}(\widehat{\phi}_{n1}(X) \ne \phi^{\textrm{Bayes}}(X)|X) -  \{{\mathbb P}_{{\cal D}_1}(\widehat{\phi}_{n1}(X) \ne \phi^{\textrm{Bayes}}(X)|X)\}^2] \\
&\le & 2 {\mathbb E} [ \ind{\widehat{\phi}_{n1}(X) \ne \phi^{\textrm{Bayes}}(X)} ],
\end{eqnarray*}
where the last equality is due to the fact that ${\cal D}_1$ and ${\cal D}_2$ are independently and identically distributed. For ease of notation, we will denote $\widehat{\phi}_{n1}$ as $\widehat{\phi}_{n}$ from now on. We further have
\begin{eqnarray*}
\textrm{CIS}(\Psi) &\le & 2 \sum_{j=0}^\infty {\mathbb E} [ \ind{\widehat{\phi}_{n}(X) \ne \phi^{\textrm{Bayes}}(X)} \ind{X\in A_j}]\\
&\le& 2 {\mathbb P}_X(0<|\eta(X)-1/2|\le \delta) + 2 \sum_{j\ge 1} {\mathbb E} [ \ind{\widehat{\phi}_{n}(X) \ne \phi^{\textrm{Bayes}}(X)} \ind{X\in A_j}].
\end{eqnarray*}
Given the event $\{\widehat{\phi}_{n} \ne \phi^{\textrm{Bayes}}\}\cap\{|\eta-1/2|>2^{j-1}\delta\}$, we have $|\widehat{\eta}_n - \eta| \ge 2^{j-1}\delta$. Therefore, for any $j\ge 1$, we have
\begin{eqnarray*}
&& {\mathbb E} [ \ind{\widehat{\phi}_{n}(X) \ne \phi^{\textrm{Bayes}}(X)} \ind{X\in A_j}]\\
&\le & {\mathbb E} [ \ind{|\widehat{\eta}_n(X) - \eta(X)| \ge 2^{j-1}\delta} \ind{2^{j-1}\delta < |\eta(X)-1/2|\le 2^j\delta}]\\
&\le & {\mathbb E}_X [ {\mathbb P}_{\cal D}(|\widehat{\eta}_n(X) - \eta(X)| \ge 2^{j-1}\delta|X) \ind{0 < |\eta(X)-1/2|\le 2^j\delta}]\\
&\le &  C_1 \exp(-C_2 a_n (2^{j-1}\delta)^2) {\mathbb P}_X (0 < |\eta(X)-1/2|\le 2^j\delta)\\
&\le & C_1 \exp(-C_2 a_n (2^{j-1}\delta)^2) C_0 (2^j\delta)^{\alpha},
\end{eqnarray*}
where the last inequality is due to margin assumption $(\ref{margin})$ and condition $(\ref{exponential})$.

Taking $\delta=a_n^{-1/2}$, we have
$$
\textrm{CIS}(\Psi) \le  C_0 a_n^{-\alpha/2} + C_0C_1 a_n^{-\alpha/2} \sum_{j\ge 1} 2^{\alpha j +1} e^{-C_2 4^{j-1}} \le C a_n^{-\alpha/2},
$$
for some $C>0$ depending only on $\alpha,C_0,C_1$ and $C_2$. \hfill $\blacksquare$\\

\subsection{Proof of Theorem \ref{thm:lowerCIS}} According to the proof of Theorem \ref{thm:upperCIS}, we have
\begin{eqnarray*}
\textrm{CIS}(\Psi)  = 2 \left\{ \mathbb E_X[ \mathbb P_{\cal D}(\widehat{\phi}_{n}(X) \ne \phi^{\textrm{Bayes}}(X)|X)] -  \mathbb E_X[\{\mathbb P_{\cal D}(\widehat{\phi}_{n}(X) \ne \phi^{\textrm{Bayes}}(X)|X)\}^2]\right\}.
\end{eqnarray*}

\citet{AT07} showed that when $\alpha\gamma\le d$, the set of probability distribution ${\cal P}_{\alpha,\gamma}$ contains a $(m,w,b,b')$-hypercube with $w=C_3 q^{-d}$, $m=\lfloor C_4 q^{d-\alpha\gamma}\rfloor$, $b=b'=C_5q^{-\gamma}$ and $q=\lfloor C_6n^{1/(2\gamma+d)} \rfloor$, with some constants $C_i\ge 0$ for $i=3,\ldots,6$ and $C_6\le 1$. Therefore, Lemma \ref{lemma:assouad} implies that the first part is bound, that is,
\begin{eqnarray*}
&& \sup_{ P\in {\cal P}_{\alpha,\gamma}} \mathbb E_X[ \mathbb P_{\cal D}(\widehat{\phi}_{n}(X) \ne \phi^{\textrm{Bayes}}(X)|X)]\\
&=& \sup_{P\in {\cal P}_{\alpha,\gamma}} \mathbb E_{\cal D} [\mathbb P_X(\widehat{\phi}_n(X)\ne \phi^{\textrm{Bayes}}(X))]\\
&\ge &\frac{mw}{2} [1-b\sqrt{nw}]\\
&=& (1-C_6)C_3C_4C_5 n^{-\alpha\gamma/(2\gamma+d)}.
\end{eqnarray*}

To bound the second part, we again consider the sets $A_j$ defined in Appendix \ref{sec:proofuppercis}. On the event  $\{\widehat{\phi}_{n} \ne \phi^{\textrm{Bayes}}\}\cap\{|\eta-1/2|>2^{j-1}\delta\}$, we have $|\widehat{\eta}_n - \eta| \ge 2^{j-1}\delta$. Letting $\delta = a_n^{-1/2}$ leads to
\begin{eqnarray*}
&& \mathbb E_X[\{\mathbb P_{\cal D}(\widehat{\phi}_{n}(X) \ne \phi^{\textrm{Bayes}}(X)|X)\}^2]\\
&= & \sum_{j=0}^{\infty} \mathbb E_X [ \{\mathbb P_{\cal D}(\{\widehat{\phi}_{n}(X) \ne \phi^{\textrm{Bayes}}(X)\}|X)\}^2 \ind{X\in A_j}]\\
&\le & \mathbb P_X (0 < |\eta(X)-1/2|\le \delta) +  \sum_{j=1}^{\infty} \mathbb E_X [ \{\mathbb P_{\cal D}(\{\widehat{\phi}_{n}(X) \ne \phi^{\textrm{Bayes}}(X)\}|X)\}^2 \ind{X\in A_j}]\\
&\le & \mathbb P_X (0 < |\eta(X)-1/2|\le \delta) +  \sum_{j\ge 1} C_1e^{-2C_2 4^{j-1}} \mathbb P_X (0 < |\eta(x)-1/2|\le 2^j\delta)\\
&\le &  C_0 a_n^{-\alpha/2} + C_0C_1 a_n^{-\alpha/2} \sum_{j\ge 1} 2^{\alpha j} e^{-2C_2 4^{j-1}}\\
&\le & C_7 a_n^{-\alpha/2},
\end{eqnarray*}
for some positive constant $C_7$ depending only on $\alpha, C_0, C_1, C_2$. When $a_n=n^{2\gamma/(2\gamma+d)}$, we have
$$
\mathbb E_X [ (\mathbb P_{\cal D}(\widehat{\phi}_{n}(X) \ne \phi^{\textrm{Bayes}}(X)|X))^2 ] \le C_7 n^{-\alpha\gamma/(2\gamma+d)}.
$$

By properly choosing constants $C_i$ such that $(1-C_6)C_3C_4C_5 - C_7>0$, we have
$$
\textrm{CIS}(\Psi)  \ge  2 [(1-C_6)C_3C_4C_5 - C_7] n^{-\alpha\gamma/(2\gamma+d)} \ge C' n^{-\alpha\gamma/(2\gamma+d)},
$$
for a constant $C'>0$. This concludes the proof of Theorem \ref{thm:lowerCIS}. \hfill $\blacksquare$

\subsection{Proof of Theorem \ref{thm:upperCISsnn}}
\label{sec:proofupperCISsnn}

According to our Theorem \ref{thm:upperCIS} and the proof of Theorem 1 in the supplementary of \citet{S12}, it is sufficient to show that for any $\alpha\ge 0$ and $\gamma\in (0,2]$, there exist positive constants $C_1,C_2$ such that for all $\delta>0$, $n\ge 1$ and $\bar{P}$-almost all $x$,
\begin{equation}
\sup_{P\in {\cal P}_{\alpha,\gamma}} \mathbb P_{\cal D}\Big(|S_n^*(x)-\eta(x)| \ge \delta\Big) \le C_1 \exp(-C_2n^{2\gamma/(2\gamma+d)}\delta^2). \label{ntp:exponential}
\end{equation}
where $S_n^*(x)=\sum_{i=1}^n w_{ni}^*\ind{Y_{(i)}=1}$ with the optimal weight $w_{ni}^*$ defined in Theorem \ref{thm:optimal} and $k^*\asymp n^{2\gamma/(2\gamma+d)}$.

According to Lemma \ref{alpha}, we have
\begin{eqnarray*}
\sum_{i=1}^{k^*}(w_{ni}^*)^2 = \frac{2(d+2)}{(d+4)k^*}\{1+O((k^*)^{-1})\} \le C_8 n^{-2\gamma/(2\gamma+d)},
\end{eqnarray*}
for some constant $C_8>0$.

Denote $\mu^*_n(x)={\mathbb E}\{S^*_n(x)\}$. According to the proof of Theorem 1 in the supplement of \citet{S12}, there exist $C_9,C_{10}>0$ such that for all $P\in {\cal P}_{\alpha,\gamma}$ and $x\in {\cal R}$,
\begin{eqnarray}
|\mu^*_n(x) - \eta(x)| &\le& \left|\sum_{i=1}^n w_{ni}^* \mathbb E\{\eta(X_{(i)}) - \eta_x(X_{(i)})\}\right|+\left|\sum_{i=1}^n w_{ni}^* \mathbb E\{\eta_x(X_{(i)})\} - \eta(x)\right| \nonumber\\
&\le& L\sum_{i=1}^n w_{ni}^* \mathbb E\{\|X_{(i)}-x\|^{\gamma}\} + \left|\sum_{i=1}^n w_{ni}^* \mathbb E\{\eta_x(X_{(i)})\} - \eta(x)\right|\nonumber\\
&\le & C_9 \sum_{i=1}^n w_{ni}^*\Big(\frac{i}{n}\Big)^{\gamma/d}\nonumber\\
&\le & C_{10} n^{-\gamma/(2\gamma+d)}. \label{halfdelta}
\end{eqnarray}

The Hoeffding's inequality says that if $Z_1,\ldots,Z_n$ are independent and $Z_i\in [a_i,b_i]$ almost surely, then we have
$$
\mathbb P\left(\Big|\sum_{i=1}^n Z_i-\mathbb E\Big[\sum_{i=1}^n Z_i\Big]\Big|\ge t\right) \le 2\exp\Big(-\frac{2t^2}{\sum_{i=1}^n(b_i-a_i)^2}\Big).
$$
Let $Z_i=w^*_{ni}\ind{Y_{(i)}=1}$ with $a_i=0$ and $b_i=w^*_{ni}$. According to $(\ref{halfdelta})$, we have that for $\delta\ge 2C_{10} n^{-\gamma/(2\gamma+d)}$ and for $\bar{P}$-almost all $x$,
\begin{eqnarray*}
\sup_{P\in {\cal P}_{\alpha,\gamma}} \mathbb P_{\cal D}\Big(|S_n^*(x)-\eta(x)| \ge \delta\Big) &\le&  \sup_{P\in {\cal P}_{\alpha,\gamma}} \mathbb P_{\cal D}\Big(|S_n^*(x)-\mu^*_n(x)| \ge \delta/2 \Big)\\
&\le & 2 \exp \{-n^{2\gamma/(2\gamma+d)}\delta^2/(2C_8)\},
\end{eqnarray*}
which implies $(\ref{ntp:exponential})$ directly. \hfill $\blacksquare$

\spacingset{1}

\newpage

\renewcommand{\theequation}{S.\arabic{equation}}
\renewcommand{\thetable}{S\arabic{table}}
\renewcommand{\thefigure}{S\arabic{figure}}
\renewcommand{\thelemma}{S.\arabic{lemma}}
\renewcommand{\thesubsection}{S.\Roman{subsection}}
\setcounter{equation}{0}
\setcounter{table}{0}
\setcounter{page}{1}
\setcounter{lemma}{0}
\setcounter{subsection}{0}

\begin{center}
\Large\bf Supplementary Materials
\end{center}

\if0\blind
{
\begin{center}
    Wei Sun, Xingye Qiao and Guang Cheng
\end{center}
} \fi

In this supplementary note, we provide lemmas for proving Theorems \ref{thm:CIS}-\ref{thm:lowerCIS}, the proofs of Corollaries \ref{cor:snn_ownn}-\ref{cor:snn_ownn_ratio}, and the calculation of $B_1$.

\subsection{A Lemma for Proving Theorem \ref{thm:CIS} (Asymptotic Equivalent Form of CIS)}
\begin{lemma}
\label{lemma}
For any distribution function $G$, constant $a$, and constant $b>0$, we have
\begin{eqnarray}
\int_{-\infty}^{\infty} \left\{G(-bu- a ) - \ind{u<0}\right\} du &=& -\frac{1}{b}\left\{ a + \int_{-\infty}^{\infty} t dG(t) \right\}, \nonumber\\
\int_{-\infty}^{\infty} u \left\{G(-bu- a ) - \ind{u<0}\right\} du &=& \frac{1}{b^2}\left\{ \frac{1}{2}a^2 + \frac{1}{2}\int_{-\infty}^{\infty} t^2dG(t) + a\int_{-\infty}^{\infty} t dG(t) \right\}. \nonumber
\end{eqnarray}
\end{lemma}

\noindent {Proof of Lemma \ref{lemma}:} We show the second equality. The proof of the first equality is similar. Note
\begin{eqnarray}
&&\int_{-\infty}^{\infty} u \left\{G(-bu- a ) - \ind{u<0}\right\} du \nonumber\\
&=& \int_{-\infty}^{0} u \left\{G(-bu- a ) - 1\right\} du + \int_{0}^{\infty} u G(-bu- a ) du \label{eq1}
\end{eqnarray}
After substitute $t=-bu- a$ for each term, we have
\begin{eqnarray*}
\int_{-\infty}^{0} u \left\{G(-bu- a ) - 1\right\} du &=& \frac{1}{b^2}\int_{-a}^{\infty} (t+a)(1-G(t))dt\\
\int_{0}^{\infty} u G(-bu- a ) du &=& \frac{1}{b^2}\int_{-\infty}^{-a} (t+a)(-G(t))dt
\end{eqnarray*}
Plugging these two into $(\ref{eq1})$, we have
\begin{eqnarray*}
&&\int_{-\infty}^{\infty} u \left\{G(-bu- a ) - \ind{u<0}\right\} du \\
&=& \frac{1}{b^2} \left\{ -\int_{-\infty}^{-a} tG(t)dt - a\int_{-\infty}^{-a} G(t)dt + \int_{-a}^{\infty} t(1-G(t))dt + a\int_{-a}^{\infty} (1-G(t))dt  \right\}\\
&=& \frac{1}{b^2} \left\{I + II + III + IV\right\}.
\end{eqnarray*}
Applying integration by part, we can calculate
\begin{eqnarray*}
I &=& -\frac{1}{2}\Big[a^2G(-a)-\int_{-\infty}^{-a} t^2 d G(t)\Big]\\
II &=& a\Big[aG(-a) + \int_{-\infty}^{-a} t d G(t)\Big] \\
III &=& \frac{1}{2}\Big[-a^2(1-G(-a)) + \int_{-a}^{\infty} t^2 d G(t)\Big] \\
IV &=& a\Big[a(1-G(-a))+\int_{-a}^{\infty} t d G(t)\Big]
\end{eqnarray*}
Plugging I-IV into $(\ref{eq1})$ leads to desirable equality. This concludes the proof of Lemma \ref{lemma}. \hfill $\blacksquare$

\subsection{A Lemma for Proving Theorem \ref{thm:optimal} (Optimal Weight)}

\begin{lemma}
\label{alpha}
Given $\alpha_i=i^{1+2/d}-(i-1)^{1+2/d}$, we have
\begin{eqnarray}
&&(1+\frac{2}{d})(i-1)^{\frac{2}{d}}\le \alpha_i \le (1+\frac{2}{d})i^{\frac{2}{d}}, \label{ineq}\\
&&\sum_{j=1}^k \alpha_j^2 = \frac{(d+2)^2}{d(d+4)}k^{1+4/d} \left\{1+O(\frac{1}{k})\right\}. \label{sum}
\end{eqnarray}
\end{lemma}

\noindent{Proof of Lemma \ref{alpha}:} First, $(\ref{ineq})$ is a direct result from the following two inequalities.
$$
(1-\frac{1}{i})^{2/d}\ge 1-\frac{2}{(i-1)d} ~~\textrm{and}~~ (1+\frac{1}{i-1})^{2/d}\ge 1+\frac{2}{id},
$$
where $i$ and $d$ are positive integers. These two inequalities hold because both differences $(1-\frac{1}{i})^{2/d}-(1-\frac{2}{(i-1)d})$ and $ (1+\frac{1}{i-1})^{2/d}- (1+\frac{2}{id})$ are decreasing in $i$ and the limit equals $0$.

Second, $(\ref{sum})$ is due to $(\ref{ineq})$ and Faulhaber's formula $\sum_{i=1}^k i^p=\frac{1}{p+1}k^{p+1} + O(k^p)$. According to $(\ref{ineq})$, we have
$$
(1+\frac{2}{d})^2\sum_{i=1}^k (i-1)^{4/d} \le \sum_{j=1}^k \alpha_j^2 \le (1+\frac{2}{d})^2\sum_{i=1}^k i^{4/d}.
$$
Due to Faulhaber's formula, $\sum_{i=1}^k i^{4/d}=\frac{d}{d+4}k^{1+4/d} +O(k^{4/d})$ and $\sum_{i=1}^k (i-1)^{4/d}=\frac{d}{d+4}k^{1+4/d} +O(k^{4/d})$, which leads to $(\ref{sum})$. This concludes the proof of Lemma \ref{alpha}. \hfill $\blacksquare$

\subsection{Strong density assumption for proving Theorems \ref{thm:upperCIS}-\ref{thm:lowerCIS}}
\label{sec:strongdensity}

The marginal distribution $\bar P$ is said to satisfy the \textit{strong density assumption} if
\begin{itemize}
\item for a compact set ${\cal R}\subset \mathbb R^d$ and constants $c_0, r_0>0$, $\bar P$ is supported on a compact $(c_0, r_0)$-regular set $A\subset {\cal R}$ satisfying $\nu_d(A\cap B_r(x)) \ge c_0 \nu_d(B_r(x))$ for all $r\in [0,r_0]$ and all $x\in A$, where $\nu_d$ denotes the $d$-dimensional Lebesgue measure and $B_r(x)$ is a closed Euclidean ball in $\mathbb R^d$ centered at $x$ and of radius $r>0$;
\item for all $x\in A$, the Lebesgue density $\bar{f}$ of $\bar P$ satisfies $\bar{f}_{\textrm{min}}\le \bar{f}(x) \le \bar{f}_{\textrm{max}}$ for some $0<\bar{f}_{\textrm{min}}<\bar{f}_{\textrm{max}}$, and $\bar{f}(x)=0$ otherwise. In addition, $\bar{f} \in \Sigma(\gamma-1,L,A).$ \hfill $\blacksquare$
\end{itemize}

\subsection{A Lemma for proving Theorem \ref{thm:lowerCIS} (Lower Bound of CIS)}

We adapt the Assouad's lemma to prove the lower bound of CIS. This lemma is of independent interest.

We first introduce an important definition called $(m,w,b,b')$-hypercube that is slightly modified from \citet{A04}. We observe independently and identically distributed training samples ${\cal D} = \{(X_i,Y_i), i=1,\ldots,n\}$ with $X_i\in {\cal X} = \cal R$ and $Y_i\in {\cal Y} = \{1,2\}$. Let ${\cal F}(\cal X,\cal Y)$ denote the set of all measurable functions mapping from $\cal X$ into $\cal Y$. Let $\cal Z = \cal X \times \cal Y$. For the distribution function $P$, we denote its corresponding probability and expectation as $\mathbb P$ and $\mathbb E$, respectively.

\begin{defi}
\citep{A04} Let $m$ be a positive integer, $w\in [0,1]$, $b\in (0,1]$ and $b'\in (0,1]$. Define the $(m,w,b,b')$-hypercube ${\cal H} = \{P_{\vec \sigma}: \vec \sigma \mydef (\sigma_1,\ldots,\sigma_m) \in \{-1,+1\}^m\}$ of probability distributions $P_{\vec \sigma}$ of $(X,Y)$ on ${\cal Z}$ as follows.

For any $P_{\vec \sigma}\in {\cal H}$, the marginal distribution of $X$ does not depend on $\vec \sigma$ and satisfies the following conditions. There exists a partition ${\cal X}_0, \ldots, {\cal X}_m$ of $\cal X$ satisfying,\\
(i) for any $j\in \{1,\ldots,m\}$, $\mathbb P_{X}(X\in {\cal X}_j) = w$;\\
(ii) for any $j\in \{0,\ldots,m\}$ and any $X\in {\cal X}_j$, we have
$$
\mathbb P_{\vec \sigma}(Y=1|X) = \frac{1+\sigma_j\psi(X)}{2}
$$
with $\sigma_0 = 1$ and $\psi: {\cal X}\rightarrow (0,1]$ satisfies for any $j\in \{1,\ldots,m\}$,
\begin{eqnarray*}
b &\mydef& \left[1- \Big(\mathbb E_{\vec \sigma}[\sqrt{1-\psi^2(X)}|X\in {\cal X}_j]\Big)^2 \right]^{1/2},\\
b' &\mydef& \mathbb E_{\vec \sigma}[\psi(X)|X\in {\cal X}_j].
\end{eqnarray*}
\end{defi}

\begin{lemma}
\label{lemma:assouad}
If a collection of probability distributions $\cal P$ contains a $(m,w,b,b')$-hypercube, then for any measurable estimator $\widehat{\phi}_n$ obtained by applying $\Psi$ to the training sample $\cal D$, we have
\begin{equation}
\sup_{P\in {\cal P}} \mathbb E^{\otimes n} [\mathbb P_X(\widehat{\phi}_n(X)\ne \phi^{\textrm{Bayes}}(X))] \ge \frac{mw}{2}[1-b\sqrt{nw}]. \label{assouad}
\end{equation}
where $\mathbb E^{\otimes n}$ is the expectation with respect to $P^{\otimes n}$.
\end{lemma}

\noindent{Proof of Lemma \ref{lemma:assouad}:} Let $\vec \sigma_{j,r} \mydef (\sigma_1,\ldots,\sigma_{j-1},r,\sigma_{j+1},\ldots,\sigma_m)$ for any $r\in \{-1,0,+1\}$. The distribution $P_{\vec \sigma_{j,0}}$ satisfies $\mathbb P_{\vec \sigma_{j,0}}(dX) = \mathbb P_{X}(dX)$, $\mathbb P_{\vec \sigma_{j,0}}(Y=1|X) = 1/2$ for any $X\in {\cal X}_j$ and $\mathbb P_{\vec \sigma_{j,0}}(Y=1|X)=\mathbb P_{\vec \sigma}(Y=1|X)$ otherwise. Let $\nu$ denote the distribution of a Rademacher variable $\sigma$ such that $\nu(\sigma=+1)=\nu(\sigma=-1)=1/2$. Denote the variational distance between two probability distributions $P_1$ and $P_2$ as
$$
V(P_1,P_2) = 1 - \int \Big(\frac{d P_1}{ d P_0} \wedge \frac{ d P_2}{ d P_0}\Big) d P_0,
$$
where $a \wedge b$ means the minimal of $a$ and $b$, and $P_1$ and $P_2$ are absolutely continuous with respect to some probability distribution $P_0$.

Lemma 5.1 in \citet{A04} showed that the variational distance between two distribution functions $P^{\otimes n}_{-1,1,\ldots,1}$ and $P^{\otimes n}_{1,1,\ldots,1}$ is bounded above. Specifically,
$$
V(P^{\otimes n}_{-1,1,\ldots,1},P^{\otimes n}_{1,1,\ldots,1}) \le b\sqrt{nw}.
$$

Note that $\cal P$ contains a $(m,w,b,b')$-hypercube and for $X\in{\cal X}_j$, $\phi^{\textrm{Bayes}}(X) = 1+\ind{\eta(X)< 1/2}=1+\ind{(1+\sigma_j\psi(X))/2 < 1/2}=(3-\sigma_j)/2$ since $\psi(X)\ne 0$. Therefore, we have
\begin{eqnarray}
&&~~~~~~~ \sup_{P\in {\cal P}} \mathbb E^{\otimes n} [\mathbb P_X(\widehat{\phi}_n(X)\ne \phi^{\textrm{Bayes}}(X))]\nonumber\\
&&\ge \sup_{\vec\sigma \in \{-1,+1\}^m} \left\{ \mathbb E^{\otimes n}_{\vec \sigma} \mathbb P_{X} (\ind{\widehat{\phi}_{n}(X)\ne \phi^{\textrm{Bayes}}(X)})   \right\} \label{asslemma_inq1}\\
&&\ge \sup_{\vec\sigma \in \{-1,+1\}^m} \left\{  \mathbb E^{\otimes n}_{\vec \sigma} \Big( \sum_{j=1}^m  \mathbb P_{X}[\ind{\widehat{\phi}_{n}(X)\ne \frac{3-\sigma_j}{2}; X\in {\cal X}_j}] \Big)\right\} \nonumber\\
&& \ge \mathbb E_{\nu^{\otimes m}} \sum_{j=1}^m   \mathbb E^{\otimes n}_{\vec \sigma} \Big(\mathbb P_{X}[\ind{\widehat{\phi}_{n}(X)\ne \frac{3-\sigma_j}{2}; X\in {\cal X}_j}] \Big)\label{asslemma_inq2} \\
&& = \mathbb E_{\nu^{\otimes m}} \sum_{j=1}^m   \mathbb E^{\otimes n}_{\vec \sigma_{j,0}} \Big( \frac{d P^{\otimes n}_{\vec \sigma}}{d P^{\otimes n}_{\vec \sigma_{j,0}}} \mathbb P_{X}[\ind{\widehat{\phi}_{n}(X)\ne \frac{3-\sigma_j}{2}; X\in {\cal X}_j}] \Big)\nonumber\\
&& = \mathbb E_{\nu^{\otimes (m-1)}(d\vec\sigma_{-j})} \sum_{j=1}^m  \mathbb E^{\otimes n}_{\vec \sigma_{j,0}} \mathbb E_{\nu(d\sigma_j)} \Big( \frac{d P^{\otimes n}_{\vec \sigma}}{d P^{\otimes n}_{\vec \sigma_{j,0}}} \mathbb P_{X}[\ind{\widehat{\phi}_{n}(X)\ne \frac{3-\sigma_j}{2}; X\in {\cal X}_j}] \Big)\label{asslemma_eq}\\
&& \ge  \mathbb E_{\nu^{\otimes (m-1)}(d\vec\sigma_{-j})} \sum_{j=1}^m \mathbb E^{\otimes n}_{\vec \sigma_{j,0}}\left[\Big(\frac{d P^{\otimes n}_{\vec \sigma_{j,-1}}}{d P^{\otimes n}_{\vec \sigma_{j,0}}} \wedge \frac{d P^{\otimes n}_{\vec \sigma_{j,+1}}}{d P^{\otimes n}_{\vec \sigma_{j,0}}}\Big) \mathbb E_{\nu(d\sigma_j)} \big(  \mathbb P_{X}[\ind{\widehat{\phi}_{n}(X)\ne \frac{3-\sigma_j}{2}; X\in {\cal X}_j}] \Big)\right]\nonumber\\\label{asslemma_inq3}\\
&& = \mathbb E_{\nu^{\otimes (m-1)}(d\vec\sigma_{-j})} \sum_{j=1}^m  \frac{1}{2} \mathbb P_{X}[\ind{X\in {\cal X}_j}] \Big[1-V(P^{\otimes n}_{\vec \sigma_{j,-1}},P^{\otimes n}_{\vec \sigma_{j,+1}}) \Big]\nonumber\\
&& = \frac{mw}{2} \Big[1-V(P^{\otimes n}_{-1,1,\ldots,1},P^{\otimes n}_{1,1,\ldots,1}) \Big]\nonumber\\
&& \ge \frac{mw}{2} [1-b\sqrt{nw}],\nonumber
\end{eqnarray}
where (\ref{asslemma_inq1}) is due to the assumption that $\cal P$ contains a $(m,w,b,b')$-hypercube, (\ref{asslemma_inq2}) is because the supremum over the $m$ Rademacher variables is no less than the corresponding expected value, (\ref{asslemma_eq}) is because we separate the space of the expectation into two parts: $\nu(d\sigma_j)$ and $\nu^{\otimes (m-1)}(d\vec\sigma_{-j})$. Finally, the inequality (\ref{asslemma_inq3}) is due to $d P^{\otimes n}_{\vec \sigma} \ge \{ d P^{\otimes n}_{\vec \sigma_{j,+1}} \wedge d P^{\otimes n}_{\vec \sigma_{j,-1}}\}$ and the latter is not random with respect to $\nu(d\sigma_j)$. This ends the proof of Lemma \ref{lemma:assouad}. \hfill $\blacksquare$\\

\subsection{Proof of Corollary \ref{cor:snn_ownn}} According to Theorems \ref{thm:upperCIS} and \ref{thm:lowerCIS}, we have, for any $\gamma\in (0,2]$,
\begin{eqnarray*}
\sup_{P\in {\cal P}_{\alpha,\gamma}} \textrm{CIS}(\textrm{SNN}) \asymp n^{-\alpha\gamma/(2\gamma+d)}.
\end{eqnarray*}

Therefore, when $\lambda\ne B_1/B_2$, we have
\begin{eqnarray*}
&&\sup_{P\in {\cal P}_{\alpha,\gamma}} \Big\{\textrm{CIS}(\textrm{SNN})-\textrm{CIS}(\textrm{OWNN})\Big\} \\
&\ge& \sup_{P\in {\cal P}_{\alpha,\gamma}} \textrm{CIS}(\textrm{SNN})- \sup_{P\in {\cal P}_{\alpha,\gamma}}\textrm{CIS}(\textrm{OWNN})\\
&\ge & C_{11} n^{-\alpha\gamma/(2\gamma+d)}.
\end{eqnarray*}
for some constant $C_{11}>0$. Here $C_{11}=0$ if and only if $\lambda = B_1/B_2$. On the other hand, we have
\begin{eqnarray*}
&&\sup_{P\in {\cal P}_{\alpha,\gamma}} \Big\{\textrm{CIS}(\textrm{SNN})-\textrm{CIS}(\textrm{OWNN})\Big\} \\
&\le& \sup_{P\in {\cal P}_{\alpha,\gamma}} \textrm{CIS}(\textrm{SNN}) + \sup_{P\in {\cal P}_{\alpha,\gamma}}\textrm{CIS}(\textrm{OWNN})\\
&\le & C_{12} n^{-\alpha\gamma/(2\gamma+d)},
\end{eqnarray*}
for some constant $C_{12}> 0$.

Furthermore, according to Theorem \ref{thm:upperCISsnn}, we have
\begin{eqnarray*}
\sup_{P\in {\cal P}_{\alpha,\gamma}} \textrm{Regret}(\textrm{SNN}) &\asymp& n^{-\gamma(1+\alpha)/(2\gamma+d)}.
\end{eqnarray*}

Similar to above arguments in CIS, we have
$$
\sup_{P\in {\cal P}_{\alpha,\gamma}} \Big\{\textrm{Regret}(\textrm{SNN})-\textrm{Regret}(\textrm{OWNN})\Big\} \asymp n^{-\gamma(1+\alpha)/(2\gamma+d)}.
$$
This concludes the proof of Corollary \ref{cor:snn_ownn}. \hfill $\blacksquare$

\subsection{Proof of Corollaries \ref{cor:cisratio} and \ref{cor:snn_ownn_ratio}} For the OWNN classifier, the optimal $k^{**}$ is a function of $k^{\textrm{opt}}$ of $k$-nearest neighbor classifier \citep{S12}. Specifically,
$$
k^{**} = \Big\lfloor \Big\{\frac{2(d+4)}{d+2}\Big\}^{\frac{d}{d+4}} k^{\textrm{opt}} \Big\rfloor.
$$
According to Theorem \ref{thm:optimal} and Lemma \ref{alpha}, we have
\begin{eqnarray*}
\sum_{i=1}^{k^*}(w_{ni}^*)^2 = \frac{2(d+2)}{(d+4)k^*}\{1+O((k^*)^{-1})\}.
\end{eqnarray*}
Therefore,
$$
\frac{\textrm{CIS}(\textrm{OWNN})}{\textrm{CIS}(\textrm{$k$NN})} \rightarrow 2^{2/(d+4)}\Big(\frac{d+2}{d+4}\Big)^{(d+2)/(d+4)}.
$$

Furthermore, for large $n$,
\begin{eqnarray*}
\frac{\textrm{CIS}(\textrm{SNN})}{\textrm{CIS}(\textrm{OWNN})} = \frac{B_3 \Big(\sum_{i=1}^{k^*} w_{ni}^{*2}\Big)^{1/2}}{B_3 \Big(\sum_{i=1}^{k^{**}} w_{ni}^{**2}\Big)^{1/2}} = \Big\{\frac{B_1}{\lambda B_2}\Big\}^{d/(2(d+4))}.
\end{eqnarray*}
The rest limit expressions in Corollaries \ref{cor:cisratio} and \ref{cor:snn_ownn_ratio} can be shown in similar manners. \hfill $\blacksquare$

\subsection{Calculation of (\ref{B1}) in Section \ref{sec:simulation}}\label{proof53}
According to the definition,
\begin{eqnarray*}
B_1 = \int_{\cal S} \frac{\bar{f}(x_0)}{4\|\dot{\eta}(x_0)\|} d \textrm{Vol}^{d-1}(x_0).
\end{eqnarray*}

When $f_1= N(0_d,\mathbb{I}_d)$ and $f_2= N(\mu,\mathbb{I}_d)$ with the prior probability $\pi_1=1/3$, we have
$$
\bar{f}(x_0)=\pi_1 f_1 + (1-\pi_1) f_2 =2(2\pi)^{-2/d} \exp\{-x_0^Tx_0/2\}/3,
$$
and
$$
\eta(x) = \frac{\pi_1 f_1}{\pi_1 f_1 + (1-\pi_1)f_2} = \Big(1+2\exp\{\mu^Tx-\mu^T\mu/2\}\Big)^{-1}.
$$

Hence, the decision boundary is
$$
{\cal S} = \{x\in {\cal R}: \eta(x)=1/2\} = \{x\in {\cal R}: 1_d^Tx=(\mu d)/2 - (\ln 2)/\mu\},
$$
where $1_d$ is a $d$-dimensional vector of all elements $1$.

Therefore, for $x_0\in {\cal S}$, we have $\dot{\eta}(x_0)=-\mu/4$ and hence
\begin{eqnarray*}
B_1 &=& \frac{2}{3\mu (2\pi)^{d/2}\sqrt{d}}\int_{\cal S} \exp\{-x_0^Tx_0/2\}  d \textrm{Vol}^{d-1}(x_0).\\
&=& \frac{\sqrt{2\pi}}{3\pi\mu d}\exp\left\{-\frac{( \mu d/2-\ln 2/\mu)^2}{2d}\right\}.
\end{eqnarray*}


\begin{thebibliography}{99}
\spacingset{1}


\bibitem[Adomavicius and Zhang, 2010]{AZ10}
Adomavicius, G. and Zhang, J. (2010). On the Stability of Recommendations Algorithms. \textit{ACM Conference on Recommender Systems}, 47--54.


\bibitem[Audibert and Tsybakov, 2007]{AT07}
Audibert, J. and Tsybakov, A. (2007). Fast Learning Rates for Plug-in Classifiers. \textit{Annals of Statistics}, \textbf{35}, 608--633.

\bibitem[Ben-Hur et al., 2002]{BEG02}
Ben-Hur, A., Elisseeff, A., and Guyon, I. (2002). A Stability Based Method for Discovering Structure in Clustered Data. \textit{Pacific Symposium on Biocomputing}, 6--17.

\bibitem[Biau et al., 2010]{BCG10}
Biau, G., C{\'e}rou, F., and Guyader, A. (2010). On the Rate of Convergence of the Bagged Nearest neighbor Estimate. \textit{Journal of Machine Learning Research}, \textbf{11}, 687--712.

\bibitem[Bjerve, 1977]{B77}
Bjerve, S. (1977). Error Bounds for Linear Combinations of Order Statistics. \textit{Annals of Statistics}, \textbf{5}, 357--369.


\bibitem[Bousquet and Elisseeff, 2002]{BE02}
Bousquet, O. and Elisseeff, A. Stability and Generalization. \textit{Journal of Machine Learning Research}, \textbf{2}, 499-526.

\bibitem[Breiman, 1996]{B96}
Breiman, L. (1996). Heuristics of Instability and Stabilization in Model Selection. \textit{Annals of Statistics}, \textbf{24}, 2350--2383.

\bibitem[B{\"u}hlmann and Yu, 2002]{BY02}
B{\"u}hlmann, P. and Yu, B. (2002). Analyzing Bagging. \textit{Annals of Statistics}, \textbf{30}, 927--961.

\bibitem[Cover and Hart, 1967]{CH67}
Cover, T. M. and Hart, P. E. (1967). Nearest Neighbor Pattern Classification. \textit{IEEE Transactions on Information Theory}, \textbf{13}, 21--27.

\bibitem[Devroye et al., 1994]{DGKL94}
Devroye, L., Gy{\"o}rfi, L., Krzyak, A. and Lugosi, G. (1994). On the Strong Universal Consistency of Nearest Neighbor Regression Function Estimates. \textit{Annals of Statistics}, \textbf{22}, 1371--1385.

\bibitem[Devroye et al., 1996]{DGL96}
Devroye, L., Gy{\"o}rfi, L., and Lugosi, G. (1996). A Probabilistic Theory of Pattern Recognition. \textit{Springer-Verlag, New York}.

\bibitem[Devroye and Wagner, 1977]{DW77}
Devroye, L. and Wagner, T. J. (1977). The Strong Uniform Consistency of Nearest Neighbor Density Estimates. \textit{Annals of Statistics}, \textbf{5}, 536--540.

\bibitem[Elisseeff et al., 2005]{EEP05}
Elisseeff, A., Evgeniou, T., and Pontil, M. (2005). Stability for Randomized Learning Algorithms. \textit{Journal of Machine Learning Research}, \textbf{6}, 55--79.

\bibitem[Fix and Hodges, 1951]{FH51}
Fix, E. and Hodges, J. L., Jr. (1951).
Discriminatory Analysis, Nonparametric Discrimination: Consistency Properties. \textit{Randolph Field, Texas}, Project 21-49-004, Report No.4.

\bibitem[Bache and Lichman, 2013]{BL13}
Bache, K. and Lichman, M. (2013). UCI Machine Learning Repository. \textit{http://archive.ics.uci.edu/ml}. Irvine, CA: University of California.


\bibitem[Gray, 2004]{G04}
Gray, A. (2004). Tubes. \textit{Progress in Mathematics}, \textbf{221}, \textit{Birkh{\"a}user, Basel.}

\bibitem[Gy{\"o}rfi, 1981]{G81}
Gy{\"o}rfi, L. (1981). The Rate of Convergence of k-NN Regression Estimates and Classification Rules. \textit{IEEE Transactions on Information Theory}, \textbf{27}, 362--364.


\bibitem[Hall et al., 2008]{HPS08}
Hall, P., Park, B., and Samworth, R. (2008). Choice of Neighbor Order in Nearest Neighbor Classification. \textit{Annals of Statistics}, \textbf{36}, 2135--2152.

\bibitem[Hall and Samworth, 2005]{HS05}
Hall, P. and Samworth, R. (2005). Properties of Bagged Nearest Neighbor Classifiers. \textit{Journal of the Royal Statistical Society, Series B}, \textbf{67}, 363--379.

\bibitem[Lee et al., 2004]{LLW04}
Lee, Y., Lin, Y., and Wahba, G. (2004). Multicategory Support Vector Machines, Theory, and Application to the Classification of Microarray Data and Satellite Radiance Data. \textit{Journal of American Statistical
Association}, \textbf{99}, 67--81.


\bibitem[Liu and Shen, 2006]{LS06}
Liu, Y. and Shen, X. (2006). Multicategory Psi-learning. \textit{Journal of American Statistical Association}, \textbf{101}, 500--509.

\bibitem[Liu and Yuan, 2011]{LY11}
Liu, Y. and Yuan, M. (2011). Reinforced Multicategory Support Vector Machines. \textit{Journal of Computational and Graphical Statistics}, \textbf{20}, 901--919.


\bibitem[Liu et al., 2010]{LRW10}
Liu, H., Roeder, K., and Wasserman, L. (2010). Stability Approach to Regularization Selection for High-Dim Graphical Models. \textit{Advances in Neural Information Processing
Systems}, \textbf{23}.



\bibitem[Meinshausen and B{\"u}hlmann, 2010]{MB10}
Meinshausen, N. and B{\"u}hlmann, P. (2010). Stability Selection. \textit{Journal of the Royal Statistical Society, Series B}, \textbf{72}, 414--473.

\bibitem[Samworth, 2012]{S12}
Samworth, R. (2012). Optimal Weighted Nearest Neighbor Classifiers. \textit{Annals of Statistics}, \textbf{40}, 2733--2763.

\bibitem[Snapp and Venkatesh, 1998]{SV98}
Snapp, R. R. and Venkatesh, S. S. (1998). Asymptotic Expansion of the K Nearest Neighbor Risk. \textit{Annals of Statistics}, \textbf{26}, 850--878.

\bibitem[Stodden et al., 2014]{SLP14}
Stodden, V. and Leisch, F. and Peng, R. (2014). Implementing reproducible research. CRC Press.




\bibitem[Stone, 1977]{S77}
Stone, C. J. (1977). Consistent Nonparametric Regression. \textit{Annals of Statistics}, \textbf{5}, 595--645.


\bibitem[Sun et al., 2013]{SWF13}
Sun, W., Wang, J., and Fang, Y. (2013). Consistent Selection of Tuning Parameters via Variable Selection Stability. \textit{Journal of Machine Learning Research}, \textbf{13}, 3419--3440.



\bibitem[Tsybakov, 2004]{T04}
Tsybakov, A. (2004). Optimal Aggregation of Classifiers in Statistical Learning. \textit{Annals of Statistics}, \textbf{32}, 135--166.

\bibitem[Wang, 2010]{W10}
Wang, J. (2010). Consistent Selection of the Number of Clusters via Cross Validation. \textit{Biometrika}, \textbf{97}, 893--904.


\bibitem[Wolberg and Mangasarian, 1990]{WM90}
Wolberg, W. H. and Mangasarian, O.L. (1990). Multisurface Method of Pattern Separation for Medical Diagnosis Applied to Breast Cytology, \textit{Proceedings of the National Academy of Sciences}, \textbf{87}, 9193--9196.



\bibitem[Yu, 2013]{Y13}
Yu, B. (2013). Stability. \textit{Bernoulli}, \textbf{19}, 1484--1500.

\end{thebibliography}

\begin{thebibliography}{99}


\bibitem[Audibert, 2004]{A04}
Audibert, J. (2004). Classification under Polynomial Entropy and Margin Assumptions and Randomized Estimators. Preprint 905, \textit{Laboratoire de Probabilites et Modeles Aleatoires, Univ. Paris VI and VII}.

\bibitem[Samworth, 2012]{S12}
Samworth, R. (2012). Optimal Weighted Nearest Neighbor Classifiers. \textit{Annals of Statistics}, \textbf{40}, 2733--2763.

\end{thebibliography}
\end{document}